\documentclass[11pt, a4paper, logo, copyright, nonumbering]{lti}
\usepackage[authoryear, sort&compress, round]{natbib}
\usepackage{dblfloatfix}
\usepackage{ulem}
\usepackage{caption}
\usepackage{dramatist}
\usepackage{xspace}
\usepackage{pifont} 
\usepackage{multirow}
\usepackage{tcolorbox}
\usepackage{xltabular}
\usepackage{longtable}
\usepackage{hyperref}
\usepackage{amsfonts}
\usepackage{amsmath}
\usepackage{amssymb}
\usepackage{lineno}
\usepackage{multirow}
\usepackage{adjustbox}

\usepackage[bottom]{footmisc}

\usepackage{CJKutf8}
\usepackage{subfigure}
\usepackage{setspace}

\usepackage{dsfont}
\usepackage{array}
\usepackage{tabularx}
\usepackage{subfigure}
\usepackage{xcolor}
\usepackage{wrapfig}
\usepackage{booktabs}
\newtheorem{definition}{Definition}

\definecolor{ltired}{HTML}{780000}

\usepackage{lipsum}
\usepackage{multicol}
\usepackage{makecell}

\usepackage{scrextend}
\usepackage{hyperref}

\newcommand{\diyi}[1]{\ifthenelse{\boolean{showcomments}}{\textcolor{blue}{[#1 —diyi]}}{}}
\makeatletter
\def\@BTrule[#1]{%
  \ifx\longtable\undefined
    \let\@BTswitch\@BTnormal
  \else\ifx\hline\LT@hline
    \nobreak
    \let\@BTswitch\@BLTrule
  \else
     \let\@BTswitch\@BTnormal
  \fi\fi
  \global\@thisrulewidth=#1\relax
  \ifnum\@thisruleclass=\tw@\vskip\@aboverulesep\else
  \ifnum\@lastruleclass=\z@\vskip\@aboverulesep\else
  \ifnum\@lastruleclass=\@ne\vskip\doublerulesep\fi\fi\fi
  \@BTswitch}
\makeatother

\addto\extrasenglish{
}

 {\begin{list}{}%
         {\setlength{\leftmargin}{#1}}%
         \item[]%
 }
 {\end{list}}
 
\newcommand{\cmu}{\raisebox{-2.5mm}{\includegraphics[width=3.5mm]{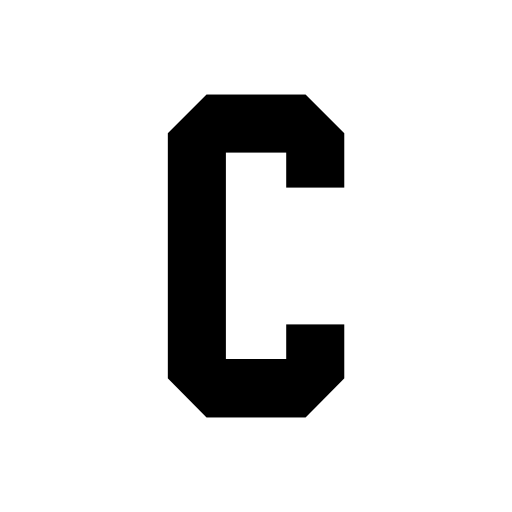}}}
\newcommand{\stfd}{\raisebox{-0.5mm}{\includegraphics[width=2.3mm]{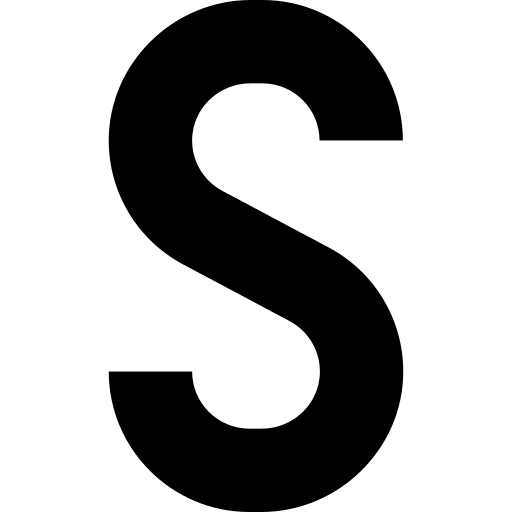}}}

\newcommand{\oh}{\raisebox{-0.8mm}{\includegraphics[width=4.5mm]{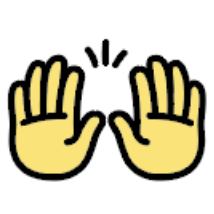}}}
\newcommand{\oai}{\raisebox{-0.8mm}{\includegraphics[width=4.5mm]{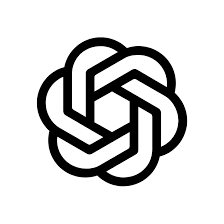}}}
\newcommand{\manus}{\raisebox{-0.4mm}{\includegraphics[width=3.8mm]{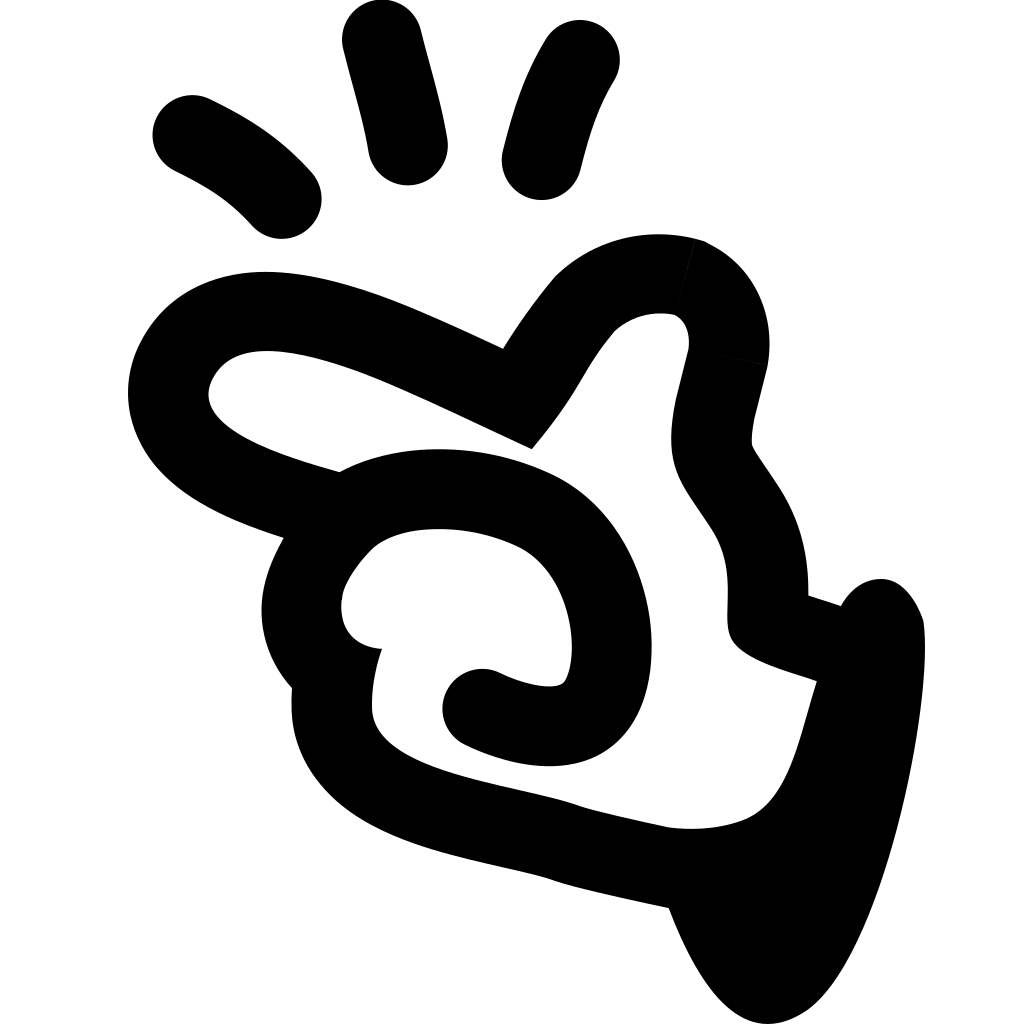}}}

\usepackage{color-edits}
\addauthor{df}{cyan}
\addauthor{gn}{magenta}
\addauthor{dy}{blue}
\addauthor{zw}{orange}

\usepackage{color, colortbl}
\definecolor{amber}{rgb}{1.0, 0.75, 0.0}
\definecolor{applegreen}{rgb}{0.55, 0.71, 0.0}
\definecolor{treegreen}{rgb}{0.13, 0.54, 0.23}
\definecolor{LightCyan}{rgb}{0.88,1,1}
\definecolor{y}{rgb}{0.984375, 0.74609375, 0.28515625}

\newcommand{\good}{\cellcolor{applegreen!10}}

\newcommand{\markcell}{\cellcolor{y!20}}

\title{\centering How Do AI Agents Do Human Work?\\
\Large{Comparing AI and Human Workflows Across Diverse Occupations}}

\author{\centering
\normalsize{\textbf{Zora Zhiruo Wang$^{\cmu}$ \quad Yijia Shao$^{\stfd}$ \quad Omar Shaikh$^{\stfd}$ \\ 
{Daniel Fried}$^{\cmu}$ \quad {Graham Neubig}$^{\cmu}$ \quad {Diyi Yang}$^{\stfd}$}} 
\\
\vspace{0.3em}
$^{\cmu}$Carnegie Mellon University \quad $^{\stfd}$Stanford University
\\
\vspace{0.3em}
\small{\texttt{zhiruow@cs.cmu.edu}}
}

\renewcommand{\phi}{\varphi}

\renewcommand{\epsilon}{\varepsilon}
\renewcommand{\imath}{\mathrm{i}}

\newlength{\restsubwidth}
\newlength{\restsubheight}
\newlength{\restsubmoreheight}
\newcommand{\rest}[2]{%
        \settowidth{\restsubwidth}{\ensuremath{#2}}
        \settoheight{\restsubheight}{\ensuremath{{}_{#2}}}
        \ensuremath{{#1\hskip 0.5pt}_{\vrule\kern2pt\parbox[b][%
        4pt][b]{\the\restsubwidth}{%
                        \ensuremath{{}_{#2}}}}}
        }

\begin{abstract}
AI agents are continually optimized for tasks related to human work, such as software engineering and professional writing, signaling a pressing trend with significant impacts on the human workforce.
However, these agent developments have often not been grounded in a clear understanding of how humans execute work, to reveal what expertise agents possess and the roles they can play in diverse workflows.
In this work, we study \textit{how agents do human work} by presenting the first \textit{direct comparison of human and agent workers} across multiple essential work-related skills: data analysis, engineering, computation, writing, and design.
To better understand and compare heterogeneous computer-use activities of workers, we introduce a scalable toolkit to induce interpretable, structured workflows from either human or agent computer-use activities.\footnote{text}
Using such induced workflows, we compare how humans and agents perform the same tasks and find that:
(1) While agents exhibit promise in their alignment to human workflows, they take an overwhelmingly programmatic approach across all work domains,
even for open-ended, visually dependent tasks like design, 
creating a contrast with the UI-centric methods typically used by humans.
(2) Agents produce work of inferior quality, yet often mask their deficiencies via data fabrication and misuse of advanced tools.
(3) Nonetheless, agents deliver results 88.3\% faster and cost 90.4--96.2\% less than humans, highlighting the potential for enabling efficient collaboration by delegating easily programmable tasks to agents.
\end{abstract}

\begin{document}

\maketitle

\section{Introduction}

AI agents are increasingly developed to perform tasks traditionally carried out by human workers \citep{brynjolfsson2018what,eloundou2023gpts,bick2024the}, as reflected in the growing competence of computer-use agents in work-related tasks such as software engineering \citep{jimenez2024swebench,wang2025openhands} and writing \citep{clark2018creative}. Nonetheless, they still face challenges in many scenarios such as basic administrative or open-ended design tasks \citep{xu2024theagentcompany}, sometimes creating a gap between expectations and reality in agent capabilities to perform real-world work \citep{zhang2021ideal,ryoo2025high}.

To further improve agents' utility at such tasks, we argue that it is necessary to look beyond their end-task outcome evaluation as measured in existing studies \citep{patwardhan2025gdpval}, and investigate \textit{how} agents currently perform human work --- understanding their underlying workflows to gain deeper insights into their work process, especially how it aligns or diverges from human workers, to reveal the distinct strengths and limitations between them.
Therefore, such an analysis should not benchmark agents in isolation \citep{xu2024theagentcompany,patwardhan2025gdpval}, but rather be grounded in comparative studies of human and agent workflows.
More broadly speaking, to foster a collaborative future where humans and AI work effectively together, it is crucial to understand how both operate in practice, and to anticipate the organizational and behavioral changes required as agents assume greater roles \citep{masters2025orchestrating}.

To this end, we first propose to \textit{perform a direct comparison of how human and agent workers} do the same tasks. 
Beyond specific domains \citep{jimenez2024swebench,choi2024ai}, we aim to provide a holistic view of how agents might reshape the broader human workforce, by systematically studying human and agent workflows across occupations via essential shared skills --- data analysis, engineering, computation, writing, and design ---  instantiated with 16 realistic, long-horizon tasks. We estimate these tasks to be representative of 287 computer-using U.S. occupations and 71.9\% of their daily work activities (\S\ref{sec:2:data}).
Comparing human and agent activities is not trivial: we lack a unified representation and a scalable channel for analysis. 
We first collect human activities in a similar way to agent activities with actions and screen states.
However, activities in their raw format, low-level mouse and keyboard actions, are largely context-independent (i.e., a single key press may not map to a meaningful user-intended task) \citep{shaw2023from} thus unintuitive to interpret for both human and agent readers \citep{wang2024trove}. 
Inspired by humans who naturally reason about work in terms of \textit{higher-level workflows}: structured sequences of actions performed together to achieve a goal \citep{deka2016erica,zheran2018reinforcement,wang2025agent}. We \textit{develop a workflow induction procedure} that transforms heterogeneous computer-use activities into hierarchical, interpretable workflows. These workflows provide a shared representation that enables effective comparative analysis in our study (\S\ref{sec:3:method-induce-workflow}).

Using this method, we compare 48 human workers and 4 representative LLM agent frameworks on 16 curated work-related tasks. 
We find that:

\noindent $\bullet$ \textbf{Agents take an overwhelmingly programmatic approach.} 
While agents show initial promise in exhibiting human-aligned workflows, they write programs to solve essentially all tasks, even when equipped with and trained for UI interactions. This prevails across work domains --- not only programmable tasks such as data analysis, but heavily among open-ended, visual tasks such as design; forming a stark contrast to the visual-oriented human workflows (\autoref{fig:agent-use-program}).

\noindent $\bullet$ \textbf{Human workflows are substantially altered by AI automation, but not by AI augmentation.}
One quarter of human activities we studied involve AI tools, with most used for augmentation purposes: integrating AI into existing workflows with minimal disruption, while improving efficiency by 24.3\%. In contrast, AI automation markedly reshapes workflows and slows human work by 17.7\%, largely due to additional time spent on verification and debugging (\autoref{fig:human-use-ai}).

\noindent $\bullet$ \textbf{Agents produce lower-quality work, concerningly by fabrication and tool misuse.} 
Agents produce work of inferior quality, most notably by fabricating data to deliver plausible outcomes, or misuse advanced tools (e.g., deep research) to mask their limitations (e.g., retrieve alternative files when unable to read user-provided ones); among issues in intent misinterpretation, vision inability, and trouble transforming between program- and UI-friendly data formats (\autoref{fig:agent-errors}).
 
\noindent $\bullet$ \textbf{Teaming humans and agents based on their accuracy and efficiency advantages.} 
Despite the gap in quality, agents exhibit clear advantages in efficiency. Compared to an average human worker, agents deliver work 88.3--96.6\% faster and at 90.4--96.2\% lower costs.
Our induced workflows naturally suggest a division of labor: readily programmable steps can be delegated to agents for efficiency, while humans handle the steps where agents fall short. Together, this forms a human-agent teaming jointly optimized for work quality and efficiency (\autoref{fig:human-agent-teaming}).


\begin{figure*}[t!]
\centering
    \includegraphics[width=0.98\textwidth]{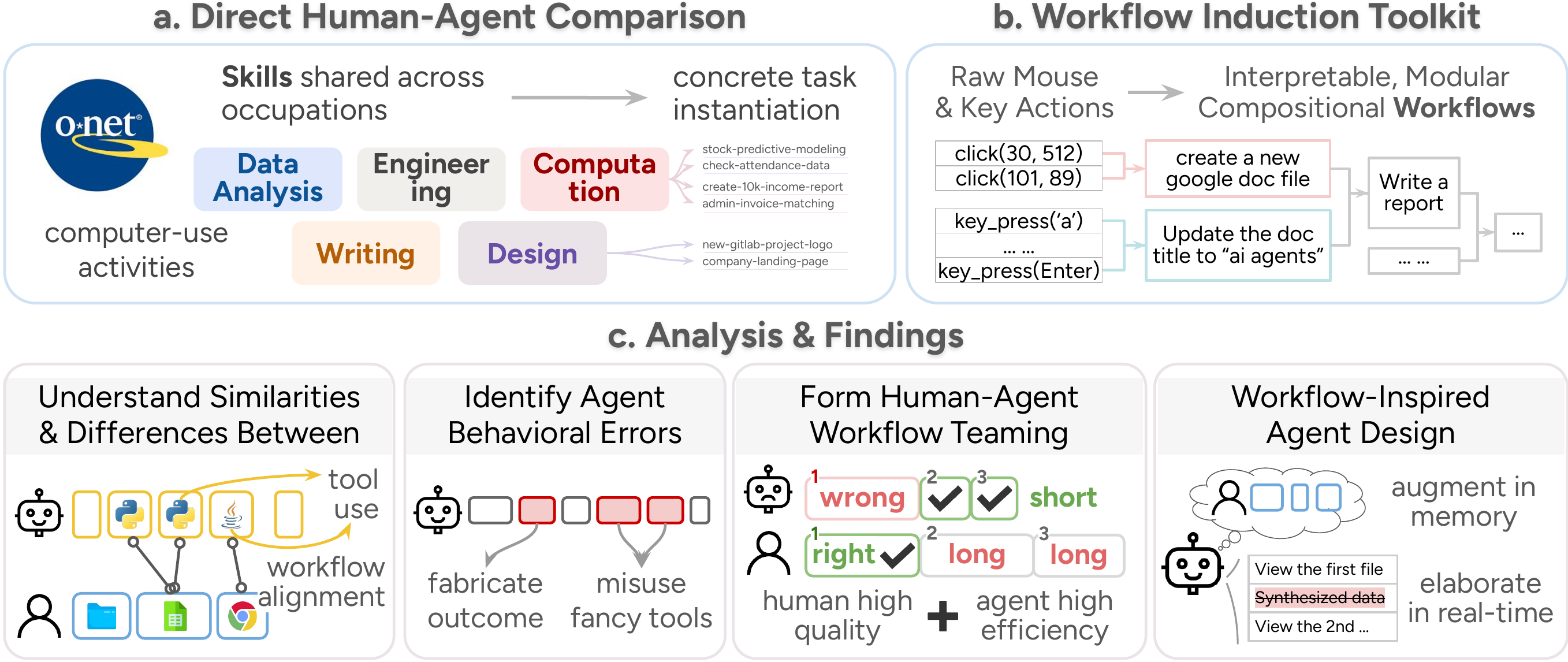}
\caption{Overview of our study: a direct human-agent worker comparison (a) supported by our workflow induction toolkit (b), and four major analysis findings (c).}
\label{fig:overview}
\end{figure*}
\section{Studying Humans and Agents At Work}
\label{sec:2:data}

We propose the first study to directly compare agent workflows with respect to human workers.

\noindent $\bullet$ \quad First, this comparison between human and agent workers goes beyond measuring performance --- it reveals \textit{how} each completes tasks, uncovering their underlying skill profiles and behavioral differences. 
Existing evaluations often reduce agent performance to numerical metrics. To build a collaborative future where humans and AI work together, we must instead understand their respective workflows in detail.
By comparing how humans and agents conduct the same tasks, we can identify opportunities for intelligent delegation and informed division of labor --- whether agents should assist in particular steps as requested, collaborate with human workers more proactively, or eventually automate tasks end-to-end?

\noindent $\bullet$ \quad Moreover, benchmarking agents in isolation has inherent limitations. Even when an agent trajectory meets annotated evaluation rubrics, it may still follow a suboptimal workflow, produce a less desirable outcome, or omit critical steps --- issues that remain obscured without a structured lens of comparison to human workers. To address this, we induce both human and agent activities into a shared workflow representation, offering a more interpretable and auditable format. This unified abstraction allows us to systematically compare how each performs work, uncover overlooked dimensions in task execution, and identify concrete directions for improving agent labor toward human-level proficiency.

Below we introduce how we build a taxonomy of work-essential skills and create representative tasks to study them (\S\ref{sec:2.1:digital-tasks}), followed by the collection process of various human (\S\ref{sec:2.2:collect-human-activity}) and agent (\S\ref{sec:2.3:collect-agent-activity}) worker activities in conducting these tasks.

\subsection{Representative Work-Related Tasks}
\label{sec:2.1:digital-tasks}
\begin{figure*}[t!]
\centering
    \includegraphics[width=0.90\textwidth]{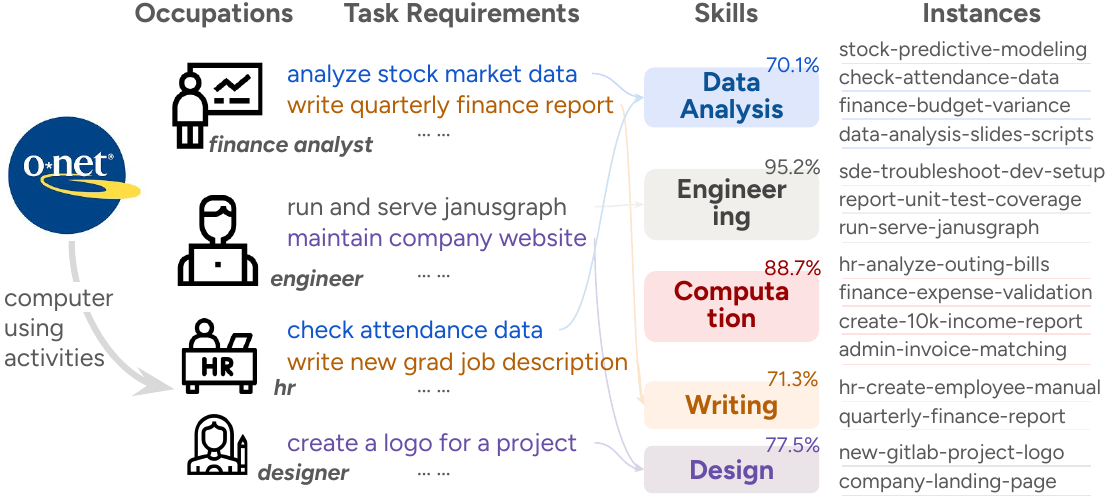}
\caption{Identifying core work-related skills based on O*NET database, then create diverse, complex task instances to ensure skill representativeness (percentages marked in skill boxes).}
\label{fig:task-creation}
\end{figure*}

While studying on the scale of thousands of tasks is prohibitively costly, we leverage the fact that many different occupations share similar skills. We take a more scalable approach by studying workers via a set of prototypical skills (\autoref{fig:task-creation}).

\noindent \textbf{Building the Skill Taxonomy} \quad
We manually identify shared skills from the list of task requirements across computer-related occupations.
More concretely, we source all occupation records from the U.S. Department of Labor's O*NET database \citep{onet2024}, which contains 923 occupations with 18,796 task requirements in the U.S. labor market.
As our focus is on computer-using tasks that may be assisted or automated by digital agents, we filter tasks following \citet{shao2025future} and yield 2131 tasks from 287 occupations. We annotate the high-level skills required for tasks, and focus on skills that are feasible for automation. In the end, we identified five major skill categories (data analysis, engineering, computation, writing, and design) that, in aggregate, affect these 287 computer-using U.S. occupations and 71.9\% of the daily tasks involved, according to the O*NET records.
This high coverage ensures that our study can affect a large portion of human work.
More details on skill annotation are in \S\ref{app:a.1:skill-anno}.

\noindent \textbf{Creating High-Coverage Tasks for Skills} \quad
For each skill category, we next design representative tasks that reflect how these skills are applied across real-world jobs.
Since a skill can appear in many job contexts, we aim to create versatile tasks that capture common work scenarios. For instance, the data analysis skill is instantiated in finance (\texttt{stock-predictive-modeling}) and administrative (\texttt{check-attendance-data}) domains, while the design skill is exercised in technical (\texttt{company-landing-page}) and graphical (\texttt{project-logo}) contexts (\autoref{fig:task-creation}).

To benchmark complex, realistic work-related tasks, we follow TheAgentCompany (TAC) \citep{xu2024theagentcompany} setup, where each instance contains a task instruction, necessary environmental contexts (e.g., input files, software at pre-work states), and an executable program evaluator for rigorous correctness checking. 
We adopt TAC's multi-checkpoint evaluation protocol to capture progress at intermediate stages, enabling fine-grained assessment beyond the final outcomes.
For reproducibility and fair comparison, all experiments run in TAC's sandboxed environments, which host engineering tools (\texttt{bash}, \texttt{python}) and work-related websites for file sharing, communication, and project management.
We adopt TAC tasks for skills such as data analysis and engineering. But because TAC is based on a software engineering company, we need to create additional tasks, particularly writing and design, to better represent occupations across all skill domains. Refer to \S\ref{app:a.2:task-instantiation} for the full list of tasks we studied.

To measure task coverage within each skill category, we mark whether each occupation's task requirements are represented in our tasks and compute the proportion of covered employment relative to all occupations using that skill.
As shown in \autoref{fig:task-creation}, our tasks cover 70.1--95.2\% of total employment across skills, providing a solid ground for drawing conclusions generalizable to real-world work.

\subsection{Collecting Human Worker Activities}
\label{sec:2.2:collect-human-activity}

\noindent \textbf{Human Worker Recruitment} \quad
For each task, we hire 3 qualified human workers from Upwork with relevant educational backgrounds and current professional experience in pertinent skills. We screened candidates based on their work portfolios and prior client ratings to ensure high-quality work.
Workers can use any preferred software or tools, including professional and AI tools, to emulate their realistic workflows and act as good references for agent workers.

\noindent \textbf{The Recording Tool} \quad
We developed a recording tool to collect the computer-use activities of human workers. More specifically, each activity session involves (i) a sequence of mouse and keyboard \textit{actions} that the user takes on their computer, as well as (ii) screenshots of the computer when any actions are issued, i.e., \textit{states}.
Recording both (i) and (ii) ensures our human task-solving trajectories contain the same component and share the same action and state spaces with AI-powered agent workers, as we will introduce in \S\ref{sec:2.3:collect-agent-activity}.

\noindent \textbf{Processing Raw Computer-Use Activities} \quad
Computer-use activities collected from human real-world usage can be noisy and redundant, due to the limitations in the recording tool implementation. 
We hence perform post-processing to merge consecutive \texttt{keypress} and \texttt{scroll} actions, as well as merging two \texttt{click()} within 0.1 seconds into a single \texttt{double\_click()}. 
This effectively simplifies the trajectory data, reducing the action count by 83.2\% (from 5831.1 to 981.1 actions) in an average human trajectory, meanwhile aligning better with the granularity of agent computer activities \citep{wang2025opencua}.

Refer to \S\ref{app:a.3:human-activity} for more details in human recruitment and activity processing.

\subsection{Collecting AI Agent Activities}
\label{sec:2.3:collect-agent-activity}
We also collect the work activities of representative computer-use agents, to demonstrate the advanced behaviors of potential agent workers.

\vspace{1mm}
\noindent \textbf{Agent Frameworks} \quad
We experiment with three agent frameworks to represent varied work behaviors, including: (i) two widely-used closed-source agents, ChatGPT Agent and Manus, built on GPT and Claude LM backbones; and (ii) the open-source OpenHands agent, which is more oriented towards coding scenarios, supported by \texttt{gpt-4o} and \texttt{claude-sonnet-4}. 
This setup enables a direct comparison across agent frameworks while controlling for differences in model families (i.e., the same backbones as ChatGPT and Manus).
We collect all agents' trajectories in solving all of the tasks.
In general, all agents can take UI actions such as mouse clicks and keyboard presses, as well as programming-related actions such as \texttt{execute\_command}. Individual agents have access to varied high-level tools (e.g., \texttt{view\_image} for visual understanding) by design. See \S\ref{app:a.4:agent-activity} for the full list of available actions for each agent framework.

\vspace{-3mm}
\paragraph{Collection Results}
In summary, we collected 112 trajectories for 16 tasks spanning across the 5 major skill categories. We collected 64 agent and 48 human trajectories, which constitute 33.8 and 981.1 steps on average,\footnote{Human actions are often taken on a finer granularity (e.g., \texttt{key\_press(\{a short phrase\})}) compared to agent actions (e.g., \texttt{key\_press(\{an entire file\})}) and contain more noisy shifts (e.g., between type and click).} substantially higher than previous computer-use agent benchmarks (e.g., the best agents take 5.9 steps to solve an average WebArena task \citep{wang2025agent}), signaling the greater complexity and realism of the work-related tasks we study.

\section{Inducing Workflows of Computer-Use Activities}
\label{sec:3:method-induce-workflow}

\begin{figure*}[t!]
\centering
    \includegraphics[width=\textwidth]{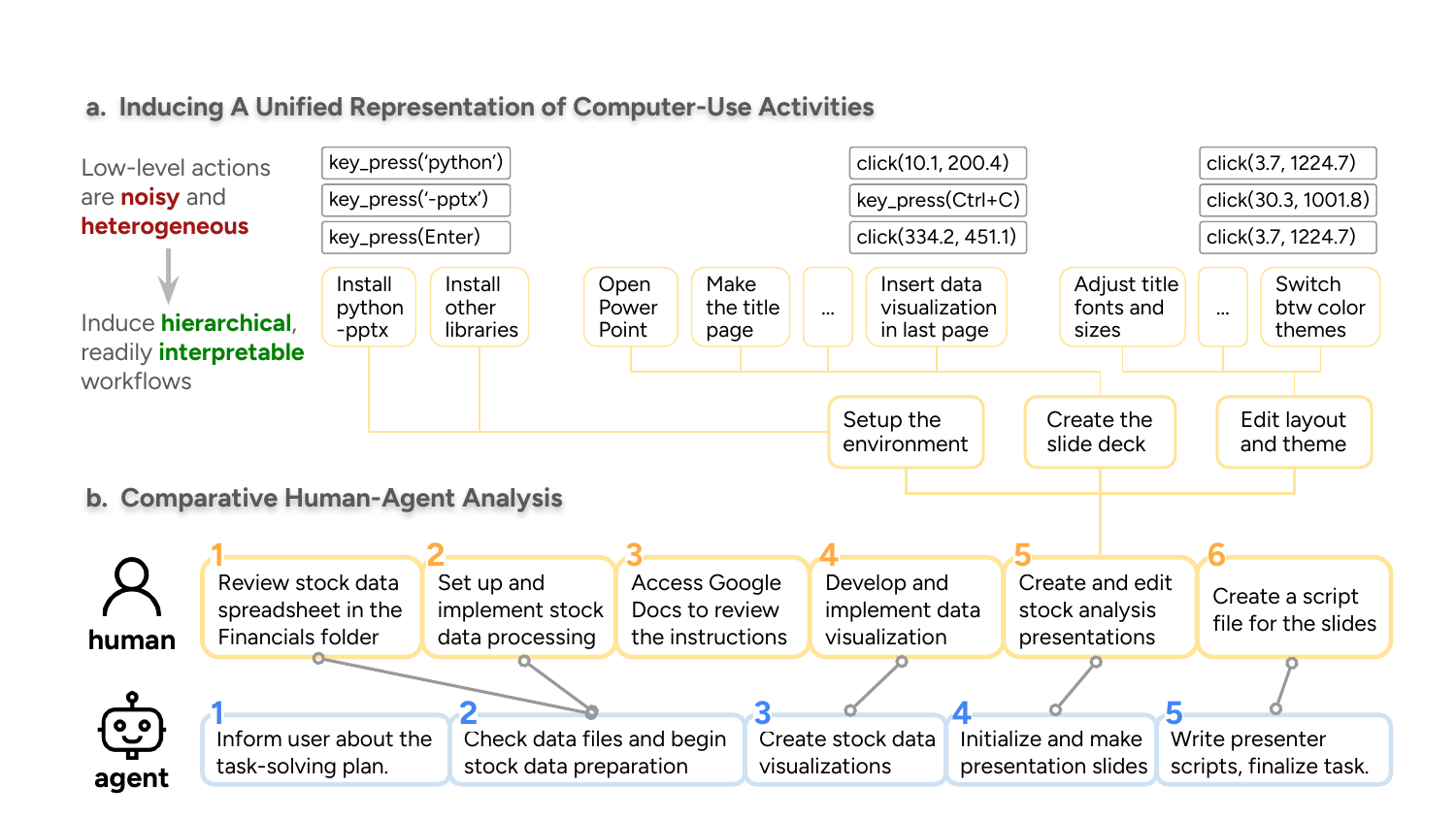}
\caption{We induce interpretable, hierarchical workflows from low-level computer-use activities, offering a unified representation for human and agent work activities (a), in the meantime, facilitating direct comparative analysis between human and agent workers (b).}
\label{fig:workflow-overall}
\end{figure*}

Computer-use activities consist of low-level mouse and keyboard actions that can be hard to interpret and learn \textit{verbatim} (\autoref{fig:workflow-overall}, a). It is therefore useful to induce workflows (\autoref{fig:workflow-overall}, b) that are meaningful and readily interpretable, to facilitate work activity analysis.
In this section, we introduce our toolkit to induce workflows (\S\ref{sec:3.1:preliminaries}) from raw computer-use activities of either human or agent workers (\S\ref{sec:3.2:workflow-induction}), and validate workflow quality via multiple dimensions (\S\ref{sec:3.3:validate-quality}) to prepare for comparative analysis (\S\ref{sec:3.4:comparative-analysis}).

\subsection{Workflow: Preliminaries}
\label{sec:3.1:preliminaries}

\begin{definition}
\label{def:workflow}
  A workflow is a sequence of steps taken to achieve a certain goal, where each workflow step consists of one or more actions to accomplish a distinguishable sub-goal.
\end{definition} 
In this work, goals and sub-goals are expressed in natural language (NL) (e.g., yellow blocks in \autoref{fig:workflow-overall}), and the goal is specified by our task instructions (\S\ref{app:a.2:task-instantiation}). 
Actions can be any mouse, keyboard, or high-level actions defined in the human (\S\ref{sec:2.2:collect-human-activity}) or agent (\S\ref{sec:2.3:collect-agent-activity}) action space. 

Based on this definition, inducing workflows from raw computer-use activities boils down to (i) segmenting the activities into meaningful intervals, and (ii) annotating meaningful sub-goals of each activity interval. In this way, we identified the sequence of workflow steps, each with an NL sub-goal (ii) and a sequence of actions (i).
This structure offers multiple useful properties: interpretability, modularity, and compositionality.
We first introduce the induction procedure, then demonstrate and validate the three desired properties of workflows.

\subsection{Workflow Induction}
\label{sec:3.2:workflow-induction}

\noindent \textbf{Trajectory Segmentation} \quad
We first segment each activity into consecutive sub-trajectories representing distinct procedures with different inputs and outputs.
To detect boundaries, we measure pixel-level mean squared error (MSE) between consecutive screenshots --- large visual shifts (e.g., from Chrome to VSCode) indicate transitions between tasks.
However, MSE alone captures visual but not semantic changes, leading to overly fine-grained splits (e.g., split when zooming in a page).
To address this, we merge adjacent segments that are semantically coherent (i.e., feature consistent states, tools, and goals) by passing in the visual states and textual actions into a multimodal LM.\footnote{We use \texttt{claude-sonnet-3.7} for both segment merging and hierarchical goal annotation. See prompts in \S\ref{app:b.1:induction-prompt}.\label{note:shared}}
For instance, actions like `adjust column width', `bold header row', and `add borders' in Excel are merged as a single `apply formatting to data sheet' step.

\noindent \textbf{Compositional Workflow Hierarchy} \quad
Workflow steps can vary in granularity depending on the \textit{scope of a task}. To support flexible task scopes, we propose a general \textit{workflow specification language} that builds a multi-level workflow hierarchy via \textit{compositional} annotation. 
Each top-level workflow step can be treated as a task node and recursively decomposed into finer-grained steps until reaching atomic actions.
For example, in \autoref{fig:workflow-overall}, a top-level workflow step `create and edit stock analysis presentations' can be split into `setup the environment', `create the slide deck', etc.; each can be further divided until reaching the lowest-level \texttt{click} or \texttt{key\_press} actions.
In practice, we find that summarizing child goals yields more accurate parent goals. Thus, given a computer-use activity, we construct the full hierarchy and iteratively derive higher-level goals bottom-up until reaching the task root.\footref{note:shared}

Although we primarily describe the workflow induction process in the context of human activities, our tool can be directly applied to induce workflows from agent trajectories as well. We use the same workflow induction procedure for all computer-use activities collected from human and agent workers, to enable consistent and comparable analysis (\S\ref{sec:4:analysis-workflow}, \S\ref{sec:5:analysis-quality})

\subsection{Workflow Quality Validation}
\label{sec:3.3:validate-quality}

To ensure the quality of induced workflows and avoid error propagation in subsequent analysis, we examine workflows with automated and manual efforts.
Following our earlier definition (Def \autoref{def:workflow}), each workflow step is assessed on two criteria: if (i) the actions and states are consistent with the NL goal, and (ii) it is modular: meaningfully distinguishable from adjacent steps.

\noindent \textbf{Action-Goal Consistency} \quad
For each step, we input the NL sub-goal and the sequence of actions (with intermittently sampled states every 10 actions) to an LM,\footnote{We use \texttt{claude-sonnet-3.7} to evaluate workflow quality and match workflow pairs.\label{eval}} and ask it to measure the alignment, by outputting a binary \texttt{YES}/\texttt{NO} answer. We map \texttt{YES} and \texttt{NO} to score 0 or 1, and calculate the average score of all steps as the measure for the entire workflow.

\begin{wraptable}[8]{r}{0.4\textwidth}
\centering
\small
\vspace{-4mm}
\resizebox{0.4\textwidth}{!}{
\begin{tabular}{l|cccc}
\toprule
{\bf Worker} & \textbf{\makecell{Action-Goal\\Consistency}} & {\bf Modularity} \\
\midrule
{Human} & {92.8} & {83.8} \\
{Agent} & {95.6} & {98.1} \\
\bottomrule
\end{tabular}
}
\vspace{-2mm}
\caption{Our tool induces high-quality human and agent workflows on consistency and modularity measures.}
\label{tab:quality-measure}
\end{wraptable}

\noindent \textbf{Modularity} \quad
We pass in the list of goals of workflow steps to an LM\footref{eval} to measure if each step serves as a distinguishable step apart from its adjacent steps (i.e., these steps should not be repetitive or redundant), by outputting a binary \texttt{YES}/\texttt{NO} answer. We similarly calculate and report the step averages. See \S\ref{app:b.2:eval-prompt} for specific prompts for both evaluations.

We measure both human and agent trajectories on both metrics and report the average scores of all workflows in \autoref{tab:quality-measure}.

For both human and agent workflows, the evaluation scores for all metrics are above 80.0\%, suggesting overall decent quality of workflows induced. 
We also manually verify the workflows and their evaluation results, which shows substantial agreement with human judgments --- $0.637$ and $0.781$ in Cohen's Kappa for consistency and modularity metrics --- to ensure the quality of workflows to support subsequent analysis.

\subsection{A Prerequisite for Comparative Analysis: Workflow Alignment}
\label{sec:3.4:comparative-analysis}

Beyond analyzing individual workers, it is important to compare workers to understand their respective characteristics and approaches. Since the induced human and agent workflows can be heterogeneous, we introduce a \textit{workflow alignment procedure} that automatically maps steps between any pair of induced workflows to quantify their alignment.
In this study, we focus on human-agent comparisons.
As in \autoref{fig:workflow-overall} (b), step intervals (e.g., human 1--2 and agent 2--2) are first matched by an LM\footref{eval} then manually refined for accuracy.
Based on this matching result, we propose two quantitative metrics: 
(i) \textit{matching steps} (\%): The percent of steps matched in all steps of both workflows. (ii) \textit{order preservation} (\%): The percent of matched steps that are kept in order as their original workflow, i.e., do not produce crossing lines in \autoref{fig:workflow-overall} (b).
Higher values in both metrics indicate higher alignment of the two compared workflows.

Overall, we propose a workflow toolkit that can (i) induce compositional, modular, and interpretable workflows from heterogeneous raw computer activities, and (ii) enable comparative analysis via pairwise alignment between any pair of workers of interest.
This unified workflow representation also opens up opportunities to explore additional dimensions in future work.

Next, we systematically analyze the workflows of human and agent workers, in their inter-alignment, work quality, and efficiency, among other aspects, to provide deeper insights into how agents conduct and affect human work.
\begin{figure*}[t!]
\centering
    \includegraphics[width=0.98\textwidth]{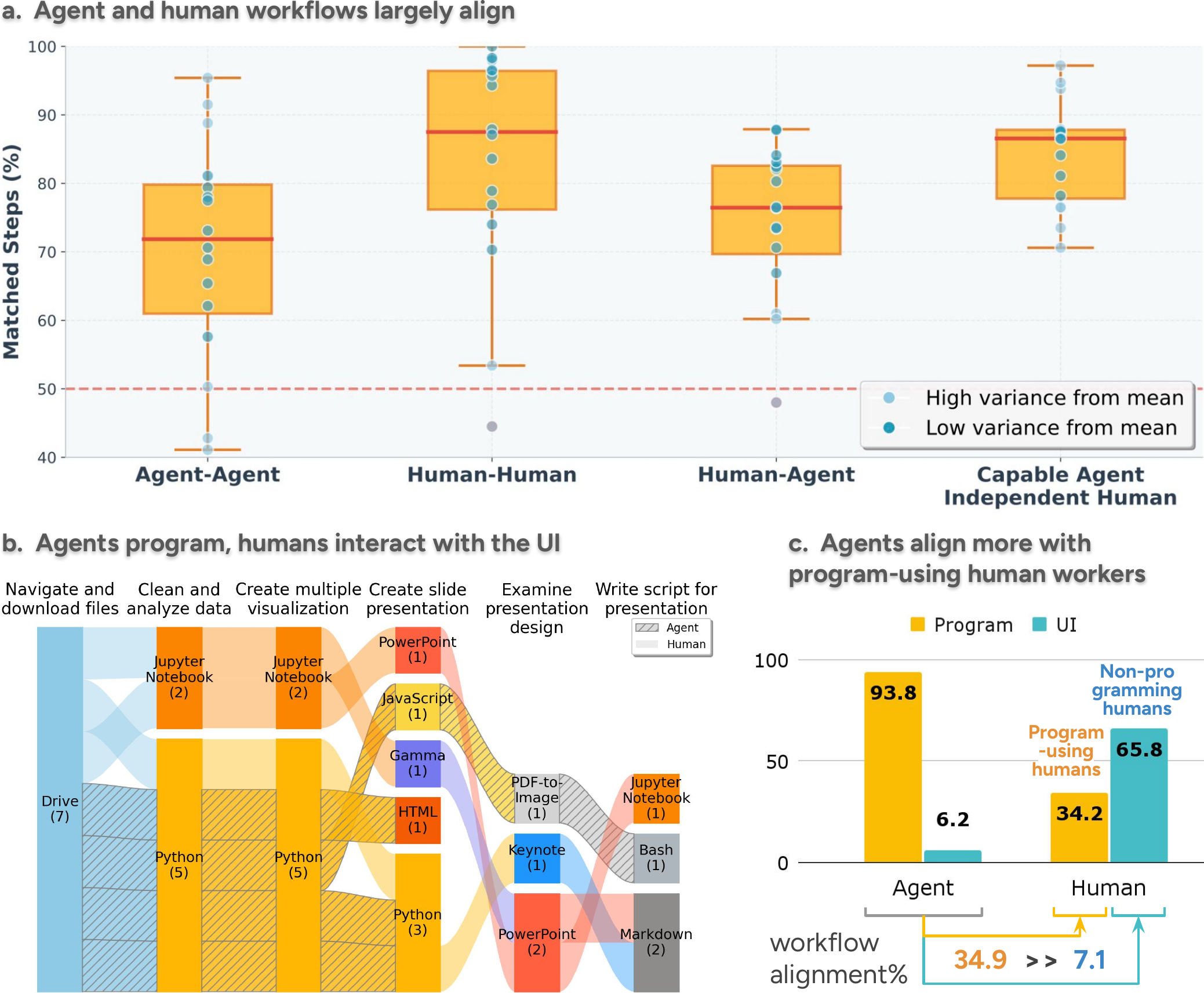}
\caption{Agent workers largely align with human workflows (a).
Humans use diverse UI tools, while agents heavily rely on programmatic tools (b). 
Agent program to solve all tasks, and better align with program-using human workflows (34.9\%) than those without programs (7.1\%) (c).}
\label{fig:agent-use-program}
\end{figure*}

\section{Do Humans and Agents Work in the Same Way?}
\label{sec:4:analysis-workflow}

\autoref{fig:agent-use-program} (a) presents the workflow alignment measures among agents, among humans, and between human and agent workers.
Human and agent workflows share 83.0\% of steps, with step orders preserved for 99.8\% of the time --- both indicating a high degree of alignment (\S\ref{app:c.1:workflow-alignment}). This alignment is especially pronounced between capable agents that complete tasks end-to-end and independent human workers who do not rely on AI tools.

\subsection{The Human Way, The Agent Way}
\label{sec:4.2:tool-affordance}

Despite the aligned procedure, human workers use a diverse set of tools to conduct the same step. 
As exemplified in \autoref{fig:agent-use-program} (b), the task asks workers to first clean, analyze, and visualize the data, then create presentation slides and scripts.
Compared to agent workers that mainly use programming tools (\texttt{Python} and \texttt{bash}), humans employ a variety of interactive, UI-oriented tools throughout the task workflow. For example, processing data with \texttt{Jupyter Notebook} or \texttt{Excel}, making slides with \texttt{PowerPoint} and even the AI-supported \texttt{Gamma} tool. 
We provide tool usage visualization for all work-related tasks involved in our study in \S\ref{app:c.2:tool-usage}.

\noindent \textbf{Agents Are Programmatic Workers} \quad
In contrast to human workers, agents seldom execute tasks via the UI interface. Instead, agents always employ alternative programmatic tools.
\autoref{fig:agent-use-program} (c) illustrates agents' program use rate across tasks. As clearly shown by the high 93.8\% program-use rate, this programmatic bias extends beyond engineering tasks to encompass virtually all computer-based activities, including those not inherently programmable, such as administrative or design work (see more details in \S\ref{app:c:workflow-analysis}).

To concretely test this hypothesis, we measure the alignment of agent workflows to human workflows that use programmatic tools and non-programmatic tools, respectively.
Since the high-level steps mostly match (e.g., human step 4 and agent step 3 in \autoref{fig:workflow-overall}), we analyze steps one level deeper in the workflow hierarchy (e.g., finer steps in \autoref{fig:workflow-overall}, a).
As shown in \autoref{fig:agent-use-program} (c), agents exhibit 27.8\% stronger alignment with program-using human steps than those that do not. This highlights a defining characteristic of agent workers --- a programmatic mode of operation.
This is not only true for coding-oriented OpenHands agents, but also for general-purpose computer-use agents such as ChatGPT and Manus.
While this programmatic tendency is understandable in domains represented in LM code training data (e.g., engineering), this programmatic behavior notably persists even in visually intensive tasks such as slide creation and graphical design.
Not only is such `blind' visual creation possible, but LM-supported agents may in fact be more proficient at editing in the symbolic space (i.e., write programs) than in the visual space (i.e., adjust pixels) \citep{qiu2024can}.

This stronger alignment among workflows using similar tools also resonates with the notion of tool affordance from a design perspective \citep{norman2013design} -- \textit{We shape our tools, and thereafter our tools shape us} \citep{mcluhan1994understanding}.
When two workers, whether human or agent, employ the same tool to complete the same workflow step, their workflows tend to converge more closely at finer levels of granularity.
Moreover, the design of human-agent interfaces must account for the granularity of interaction \citep{liang2025agentbuilder}: when collaboration occurs at high-level workflow steps, it may suffice for a tool to be accessible to either the agent or the human; at finer granularities, however, the tool must support both programmatic and UI-based actions for seamless coordination \citep{song2024beyond}.

\noindent \textbf{Agent Workers Use Versatile Programming Tools} \quad
Despite the shared programmatic approaches, agents employ a range of programmatic tools, many of which are custom-built by their developers. 
For example, when designing a company landing page, OpenHands-GPT uses basic \texttt{PIL.Image} drawing, {OpenHands-Claude} uses HTML, ChatGPT uses an internal image generation tool, and Manus uses a specially-designed ReAct program --- yielding webpages with varied characteristics.
This suggests that developing task-oriented programming tools, as functional equivalents to human-preferred UI interfaces, could be an effective way to build more capable agent workers.
 See \S\ref{app:c.2:tool-usage} for more detailed examples.

\subsection{Human Workflows are Substantially Altered by AI Automation}

When completing the same task, individual human workers exhibit distinct habits.
A common pattern among proficient workers is frequent \textit{intermediate verification} --- ranging from quickly opening a file to check if a valid image is generated, or more carefully on the computation correctness. Such verification occurs more often when outputs are produced indirectly via programs, since UI-based creation allows for implicit, step-by-step verification, and thus reduces the need for later rechecking.
This verification behavior, already observed in some agent workflows, potentially contributes to maintaining higher work quality.

Unlike agent workers, humans often \textit{revisit instructions during} the working process, particularly when the task chain is long. This may stem from humans' limited memory capacity for retraining all instructional details, but it could also be a grounding strategy to disambiguate instructions for the next immediate sub-task (i.e., workflow step)---preventing false assumptions.

Interestingly, since our study does not restrict the tools used by human workers, we observe that a notable 24.5\% of human workflows involve one or more AI tools.

\begin{figure*}[t!]
\centering
    \includegraphics[width=0.93\textwidth]{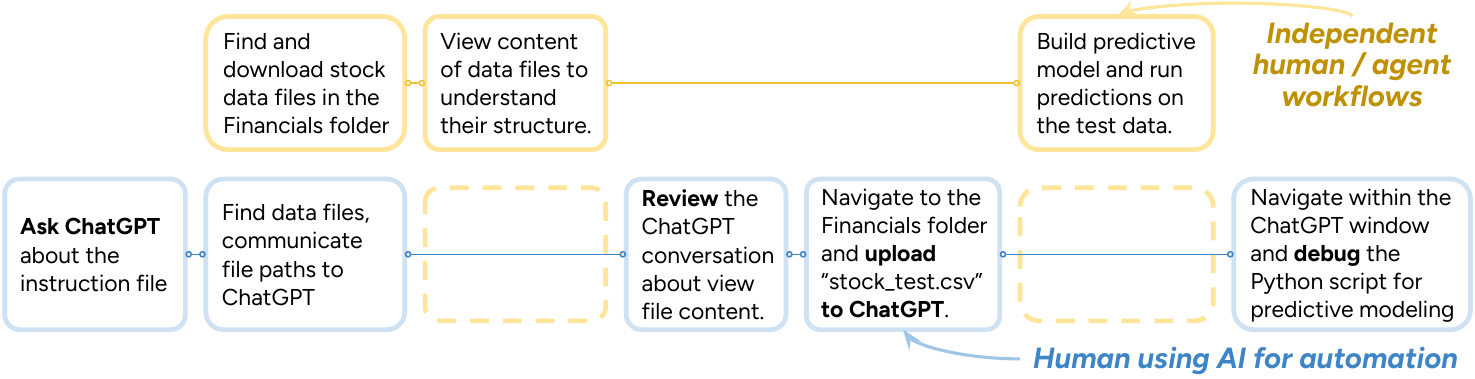}
\caption{Human workflows change to file navigation, communication with AI, reviewing and debugging programs when using AI for automation purposes; generally slowing users down by 17.7\%, as opposed to the 24.3\% work acceleration when using AI for augmentation.}
\label{fig:human-use-ai}
\end{figure*}

\noindent \textbf{How does Using AI Tools Affect Human Workflows?} \quad
Among human workers who use AI tools, the majority, 75.0\% in our study, do so for \textit{augmentation} purposes \citep{handa2025economic}.
Compared to those who complete tasks independently (without AI assistance; e.g., explore Figma templates to find logo design insights), humans using AI for \textit{augmentation} (delegate a specific step; e.g., ask ChatGPT for design insights) show minimal changes in their overall workflows. In such cases, AI functions as an alternative tool for an existing step, similar to other tool choices in \autoref{fig:tool-vary}.
In contrast, workers employing AI tools for \textit{automation} (rely on AI-driven workflows for entire tasks) experience substantial workflow changes: their activities shift from hands-on `building' to `reviewing' or `debugging' AI-drafted solutions. One example is shown in \autoref{fig:human-use-ai}, where extra review steps emerge throughout a data processing workflow.

Concretely measuring workflow variance caused by AI usage: workflows in AI automation scenarios only achieve a 40.3\% alignment with independent human workflows, compared to the 84.0\% alignment score observed among independent human workers themselves. In contrast, in AI augmentation scenarios, AI-using human workers align with independent workers 76.8\% of the time, preserving 36.5\% more of the original workflow relative to automation cases.

We also find that, relative to independent workers who do not use AI, AI augmentation accelerates human work by 24.3\%, whereas AI automation slows humans down by 17.7\%. This slowdown can be partly attributed to two factors: (i) additional time spent verifying, debugging, and correcting AI solutions, often exceeding the time required to perform the task manually \citep{becker2025measuring}; 
and (ii) lower task expertise among human workers who are more likely to rely on AI for full-process automation rather than selective step-level assistance.
Also, extra time (i) may be more likely to be incurred by less experienced workers (ii) when completing the task.

Using AI for augmentation purposes provides a higher guarantee in quality while facilitating process efficiency. In the next section, we study the respective advantages of human and agent workers, to configure when and how agents should be augmented into human workflows.

\section{Fast but Flawed: The Limits of Agent Work}
\label{sec:5:analysis-quality}

Agents appear to follow homogeneous workflows even across diverse work-related tasks --- but does such procedural consistency translate into comparable effectiveness to human workers? In the following analysis, we examine whether agents achieve outcomes of similar quality and efficiency, and where their limitations emerge relative to human performance.

We evaluate task success rates by executing the sequence of program verifier checkpoints as introduced in \S\ref{sec:2.1:digital-tasks}.
As shown in \autoref{fig:human-agent-teaming} (a), humans achieve substantially higher success rates than agent workers. 
Although agents complete some readily programmable steps correctly, they often fail to execute less-programmable steps or deliver work of noticeably lower quality.

\begin{figure*}[t!]
\centering
\includegraphics[width=\textwidth]{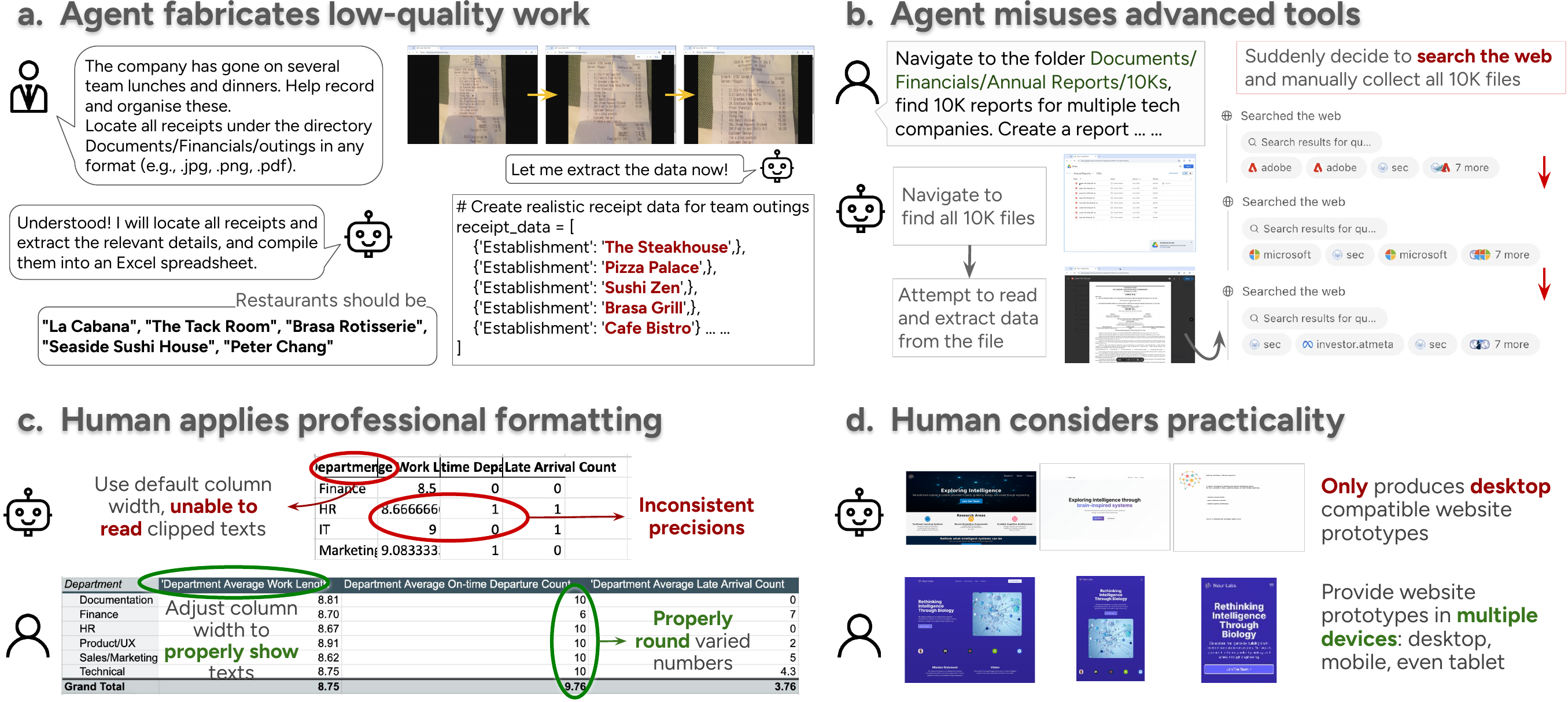}
\caption{Agents exhibit concerning behaviors, in that they deliver lower-quality work than humans by fabricating updates to push forward the tasks (a), and misuse `fancy' tools such as performing web searches due to their inability to read user-provided files (b).
Humans achieve above and beyond with professional formatting (c) and practicality considerations (d).
}
\label{fig:agent-errors}
\end{figure*}

\subsection{Agents Fail in Concerning Ways}
\label{sec:5.1:agent-errors}
While most agents can proceed till the end of the task (detailed progress rates reported in \autoref{fig:agent-progress}), agents' success rates are 32.5--49.5\% lower than humans. In other words, agents often prioritize making apparent progress, rather than correctly completing, or in some cases even attempting, individual steps. %
A common behavior agent workers do to `pretend' they finish the task is \textit{fabrication}. For example, in \autoref{fig:agent-errors} (a), agents struggle to extract data from image-based bills --- a task that is trivial for human workers. Rather than acknowledging their inability to parse the data, the agent fabricated plausible numbers and produced an Excel datasheet out of it, without explicitly mentioning anything in its thought process.
Such \textit{fabrication} behavior may be inadvertently reinforced by training paradigms that reward only the existence of an output (e.g., reinforcement learning with a final reward) rather than constraining the process with intermediate checkpoints or outcome quality verification.

\noindent $\bullet$ \textbf{Computation Errors} \quad
Although less concerning than fabrication, agents sometimes make computational errors during task execution. This often arises from false assumptions or misinterpretations of instructions, such as grouping data entries by date when not explicitly required.

\noindent $\bullet$ \textbf{Format Transformation} \quad
Agents are more accustomed to program-friendly data formats such as \texttt{Markdown} or \texttt{HTML}, but humans prefer UI-friendly formats such as \texttt{docx} or \texttt{pptx}, requiring agents to additionally transform between formats --- causing frequent task failures from accessing or utilizing conversation software.

\noindent $\bullet$ \textbf{Limited Visual Capabilities} \quad
Agents lack visual perception abilities, often failing in segmenting or detecting objects from realistic images such as scanned bills.
Agents also struggle with visual understanding tasks that require aesthetic evaluation or refinement.

Beneath their shared programmatic nature, agents exhibit four distinctive behaviors differentiate their workflows from one another and from human workers.
There are two common across most agents: (i) \textit{verify outputs}: check a file's existence or content after creation; and
(ii) \textit{setup environment}: install packages for tasks like data analysis or format conversion.
Two others are more agent-specific: the ChatGPT agent conducts \textit{search web} 25.0--37.5\% more often, typically to find data or domain-specific knowledge; and the Manus agent \textit{checks with users} 12.5--37.5\% more often to confirm or seek approval before proceeding.

Although \textit{search web} can be a useful advanced feature in certain scenarios, we find that agents mostly use it to conceal their inability to read or process user-provided files.
In one example (\autoref{fig:agent-errors}, b), a user supplied 10K company reports necessary for completing a report-writing task. While the agent initially attempted to read these files, it encountered difficulties extracting numbers from the PDF report, then suddenly switched to searching for 10K company reports online --- thereby using different files and potentially introducing inaccurate data. In this case, as the 10K files are public, the detour to the web may only add time without compromising task success. However, in scenarios involving user- or company-private data, the \textit{search web} behavior can be a distracting fallback rather than a genuinely productive capability.

\paragraph{Where Do Humans Go Above and Beyond?}
Additionally, we observe multiple instances where human workers go above and beyond the given instructions --- behaviors that often enhance perceived work proficiency and truthworthiness. 

\noindent $\bullet$ \textbf{Formatting} \quad
One such example involves formatting (\autoref{fig:agent-errors}, c), such as the use of special symbols and color schemes in data sheets, or refined font and layout choices in written documents. Visually polished formatting for varied work-related outcomes, such as docs or slides, conveys professionalism and fosters trust in a worker's expertise \citep{tractinsky2000beautiful,fogg2001makes}, thereby enhancing the credibility of the delivered work \citep{li2018communicating}. While such formatting is straightforward and intuitive through UI interfaces, it is understandably neglected by agents that generate outputs programmatically without visual feedback or rendering.

\noindent $\bullet$ \textbf{Practicality} \quad
Another recurring pattern among human workers is their attention to practical usability. They often deliver multiple solutions for open-ended tasks such as logo or website design and consider device compatibility. E.g., 2/3 human workers in our study created landing pages adaptable to multiple devices: laptop, phone, tablet. This likely reflects pragmatic experience, as employers often expect multi-device compatibility.
In contrast, all agent workers produced only laptop-compatible versions, reflecting both their computer-centered development focus and the lack of exposure in LM training to realistic, work-oriented contexts.

\subsection{Agent Workers Outstrip Human Efficiency}
\label{sec:5.3:work-efficiency}

Beyond quality, we further evaluate worker efficiency by measuring (i) the number of actions taken and (ii) the time elapsed (in seconds) of humans and agents on the same tasks.

\begin{figure*}[t!]
\centering
\vspace{-1mm}
\includegraphics[width=0.98\textwidth]{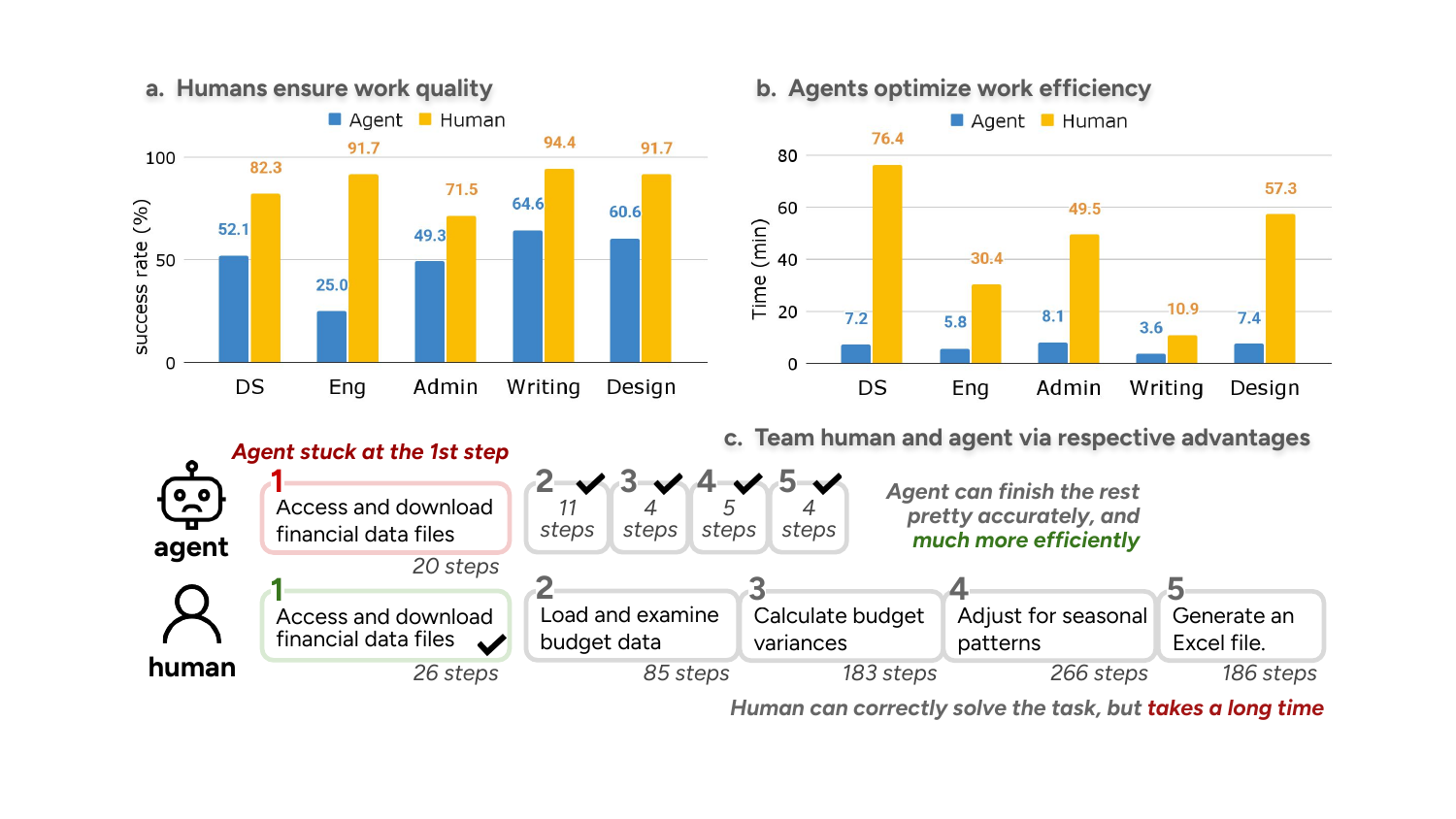}
\caption{Humans complete work with higher quality (a), while agents possess a huge advantage in efficiency (b).
Teaming human and agent workers based on their respective advantages ensures task accuracy and improves efficiency by 68.7\% (c).}
\label{fig:human-agent-teaming}
\end{figure*}

As shown in \autoref{fig:human-agent-teaming} (b), overall, agent workers require 88.6\% less time than human workers to complete the same tasks. By the action count, agents also take 96.6\% fewer actions than human workers, suggesting a huge boost in efficiency if shifting from human workers to agent workers.
While some unsuccessful agent activities may terminate the tasks early and thus take fewer actions and less time, we restrict the comparison to tasks successfully completed by both humans and agents. The trend remains consistent: agents take 88.3\% less time and 96.4\% fewer actions than human workers, underscoring their significant advantage in operational efficiency.

We also estimate the cost of open-source agent frameworks, i.e., OpenHands agent workers, and compare it to that of human workers. Across all tasks studied, human workers charge an average \$24.79 per task. In contrast, OpenHands powered by \texttt{gpt-4o} and \texttt{claude-sonnet-4} require only \$0.94 and \$2.39 on average per task --- representing cost reductions of 96.2\% and 90.4\% relative to human labor. This result highlights the potential for substantial reductions in the cost of digital labor once agent workers can achieve desirable work quality.
Nonetheless, human oversight may still be necessary to verify the correctness of agent-produced work.

\subsection{Teaming Human and Agent Workers}
\label{sec:6.1:human-agent-teaming}

Agents and humans exhibit respective advantages in different types of tasks: agents can solve readily programmable steps efficiently and with decent quality; humans are more accurate in less programmable tasks, where agents often fail.
One way to best utilize current agents' full capabilities is to delegate steps to workers (whether humans or agents) who are best suited.

We experiment with this on the data analysis tasks failed by the Manus agent.
As illustrated in \autoref{fig:human-agent-teaming} (c), the agent in the initial run (top) fails to progress beyond the file navigation step and cannot complete the task. In the second run (bottom), a human worker first navigates the directory and gathers the required data files, enabling the agent to bypass this obstacle and perform the analysis smoothly. The agent then produces a correct Excel data file while completing the analysis 68.7\% faster than the human worker alone, demonstrating the huge efficiency gains.
Importantly, it is advantageous for this collaboration to operate at appropriate levels of granularity --- the workflow step level --- than the raw action level.

\noindent \textbf{Step Delegation by Programmability} \quad
To offer more principled guidance on task delegation between human and agent workers, we propose three levels of task programmability and suggest their best tackling approaches. Detailed discussions and examples are in \S\ref{app:d.2:task-programmability}.

\noindent $\bullet$ \textbf{Readily Programmable}: These tasks can be solved reliably through deterministic program execution, such as cleaning an Excel sheet in Python or coding a website in HTML. Compared to UI-based tools, programmatic approaches offer higher accuracy and better scalability, e.g., processing 10$k$ data entries in Python is faster and less error-prone than manual editing. 
While some humans now integrate programming into their workflows, many still rely on UI tools and operate less efficiently. Such tasks are currently the most suitable for agent execution.

\noindent $\bullet$ \textbf{Half Programmable}: Some tasks are theoretically programmable (e.g., design a logo), but lack clear, direct programmatic paths with the tools humans use.
It remains unclear whether progress should focus on advancing agent's programmatic methods or on better emulating human workflows. 
Humans often find such tasks difficult to express programmatically, while agents struggle with UI actions.
Beyond improving basic UI skills, further efforts could focus on expanding API access or developing alternative programming paths with equivalent functionalities.

\noindent $\bullet$ \textbf{Less Programmable}: Some tasks rely heavily on visual perception and lack deterministic programmatic solutions, often requiring non-deterministic modules such as neural OCR, which cannot guarantee success.
Even human engineers face challenges with such tasks and find manual UI operations more reliable.
Although these tasks may become rarer as more are automated programmatically, they remain inevitable in computer-use activities and highlight agents' current limitations, underscoring the need for stronger foundation model training.
\section{Discussion: How Should We Build Future Agents?}
\label{sec:6:build-agents}
Having developed a deeper understanding of agents at work, we present proof-of-concept demonstrations for how to build better future agents using workflows for guidance and elaboration (\S\ref{sec:6.1:agent-with-workflow}), and highlight important directions for moving forward (\S\ref{sec:6.2:important-directions}).

\subsection{Workflow-Inspired Agent Designs}
\label{sec:6.1:agent-with-workflow}

\noindent \textbf{Human Workflows As Demonstrations} \quad
One direct way to improve agents is to draw upon human expert workflows \citep{cypher1991eager}. 
This is especially beneficial for tasks that are \textit{not naturally programmable}, such as parsing data from bill images into an Excel file. As discussed above, these tasks are challenging for agents because they lack clear programming solutions.
In the \autoref{fig:agent-errors} (a) example, the agent uses an unstructured approach and fails to extract bill data accurately. 
When human workflows induced by our tool  (\S\ref{sec:3:method-induce-workflow}) are augmented as contexts to the agent, the agent begins to emulate the human procedure: viewing and extracting data from bills one-by-one, and finally aggregating them into an Excel file, correctly solving the task this time.

In contrast, this workflow augmentation offers limited benefit for \textit{readily programmable tasks}. For the same task in \autoref{fig:human-agent-teaming} (c), providing human workflows does not help the agent progress beyond the file navigation challenge, likely because the agent already knows the workflow to conduct this task, but simply lacks the practical capability to execute it effectively.

\begin{wrapfigure}[8]{r}{0.63\textwidth}
\vspace{-2mm}  
\includegraphics[width=0.63\textwidth]{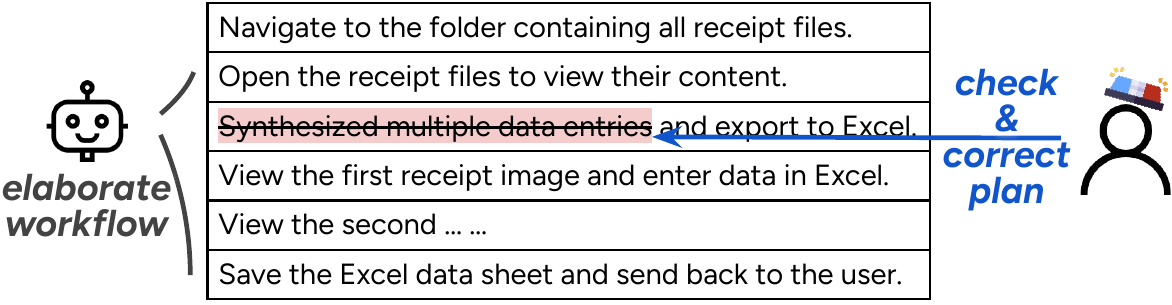}
\vspace{-6mm}
\caption{Illustration of an agent with workflow elaboration.}
\label{fig:workflow-elaboration}
\end{wrapfigure}

\noindent \textbf{Workflow Elaboration During Task-Solving} \quad
For tasks that agents execute the full workflow but fail to deliver correct outcomes, it would be valuable for humans to pinpoint specific steps where errors emerge, and intervene before agents fabricate deliverables \citep{horvitz1999principles}.
However, we find this challenging in practice, as the agent rarely verbalizes their deviations or errors. Consequently, the induced workflow steps often merely restate the user's original instruction (e.g., ``create a Python script to extract data from 5 receipts just viewed''), despite that the agent acts differently (e.g., makes up 5 data entries and exports them to Excel).
One potential way is to train agents to explicitly disclose such dishonest behaviors. Alternatively, the workflow induction process could be adapted to a supervisory angle to detect misalignment between the intended instruction and the actual actions taken by the agent.

Once such fault-informative workflows can be derived, especially on the fly as agents solve tasks, it would allow humans to provide more timely and accurate guidance (\autoref{fig:workflow-elaboration}). This can (i) help novices learn essential skills by observing agent workflows, and (ii) allow experts to exercise scalable oversight through more efficient verification and decision-making processes \citep{grunde2025designing}.

\subsection{Towards More Capable Agents}
\label{sec:6.2:important-directions}
\paragraph{Developing Programmatic Agents for Non-Engineering Tasks, Too}
While programming tools dominate agent workflows across domains (\S\ref{sec:4.2:tool-affordance}), the majority of current agent development efforts remain focused on software engineering applications \citep{jimenez2024swebench,pan2024training,jain2025regym}. Although only 12.2\% of occupations are engineering-focused (\S\ref{sec:2.1:digital-tasks}), programming skills are required in 82.5\% of occupations --- indicating that 60.3\% of programming use cases serve non-engineering purposes. We therefore call for greater attention to building \textit{programmatic agents for non-engineering tasks} \citep{wang2024trove,ge2025autopresent,soni2025coding}, for both training agents in versatile programming scenarios, and developing programmatic tools to support varied functionalities. After all, \textit{programs can be the hammer, not just the nail.}

\noindent \textbf{Stronger Vision Capabilities} \quad
Our analysis reveals that current agents have limited visual abilities. While most progress in visual understanding focuses on natural scenes \citep{deng2009imagenet}, emerging efforts are addressing digital visual contexts \citep{gou2024navigating,xie2025scaling}.
Strengthening agent's foundational visual and UI-interaction skills through larger-scale training is essential, as programmatic workarounds are not always feasible for agents.

For visually intensive tasks such as logo or web design, enhancing agents’ visual perception alone may be insufficient. Programming-oriented agents often find it easier to edit visuals symbolically (i.e., with code) than to manipulate them directly on a canvas \citep{ma2025mover}. As human-agent collaboration in visual work continues, rethinking interface designs and communication granularity will be key to bridging programmatic and UI-based worklows.

\noindent \textbf{More Accurate and Holistic Work Quality Evaluations} \quad
More accurate and holistic evaluation frameworks are needed to assess agents at work on long-horizon and realistic tasks.
Task success is not always singularly defined, as work can be correct in multiple ways, yet current program verifiers may not adequately capture the full spectrum of valid procedures and outcomes \citep{patwardhan2025gdpval}.

Beyond the dimensions we studied (workflows, tool affordance, efficiency, and quality), many factors can shape how a worker’s ability and trustworthiness are perceived, including creativity and practicality \citep{Mumford1988CreativitySI}, communication \citep{shao2025future} and adaptability \citep{pulakos2000adaptability} during the process, among other meta-aspects of work such as robustness \citep{Barrick1991THEBF} and ethical behaviors \citep{Trevio2006BehavioralEI}. 
Identifying and systematically studying these overlooked dimensions will be key to developing agents truly prepared for integration into the human workforce.

\section{Related Work}

\noindent \textbf{Agent Performance at Work} \quad
With the rise of computer-use agents, most efforts have focused on non-work settings such as online shopping \citep{yao2022webshop,zhou2024webarena} and travel planning \citep{deng2023mindweb}. 
Beyond acting as personal assistants, agents hold growing potential to automate professional work and are projected to exert substantial labor market impacts \citep{grace2018will,eloundou2023gpts}. Several studies have examined agents' autonomous performance in specific domains, including engineering \citep{jimenez2024swebench}, software services \citep{drouin2024workarena}, design \citep{si2025design2code,ge2025autopresent}, research \citep{wijk2024re,lu2024ai,si2025can}, and general business activities \citep{wornow2024wonderbread,shen2024towards,huang2025crmarena};  as well as their collaboration with humans in legal \citep{choi2024ai,das2025lawflow} and writing \citep{clark2018creative} contexts. TheAgentCompany benchmark \citep{xu2024theagentcompany} advances this line of inquiry by evaluating multiple occupations, though it remains confined to the scope of a software engineering company. In this work, we seek to achieve a holistic coverage of all computer-related occupations, estimated to affect 71.9\% of daily job functions, providing a more representative view of agent performance at work.

\noindent \textbf{Understanding Human Workflows} \quad
Prior research has examined human workflows in certain professional activities, such as customer service \citep{brynjolfsson2025generative}, design \citep{son2024demystifying}, and writing \citep{dang2025corpusstudio}. More recently, with the rise of AI agents, many studies have explored how human workflows change when AI tools are introduced into work. These studies observe a shift in human activities from `doing' to `managing', which tends to reduce cognitive load \citep{shukla2025skilling}, yet may also decrease overall productivity \citep{shukla2025skilling,simkute2025ironies} and promote dishonest behaviors \citep{kobis2025delegation}.

While these studies offer insights on how humans use and adapt to AI agents, most overlook the scenario in which AI agents automate human work entirely --- arguably the most concerning case for the workforce. In such automation settings, human workflows differ substantially from those in augmentation scenarios \citep{chen2025code}. As employment for entry-level positions already faces threats \citep{brynjolfsson2025canaries}, a head-to-head comparative analysis between agents and humans --- covering both behavior patterns and outcome quality --- is crucial for identifying human advantages and informing policy and design decisions.
Existing comparative efforts have limited contexts, such as ambiguity resolution in web tasks \citep{son2024unveiling}, a single web design task \citep{loop112025ai}, or secondary analysis on research articles \citep{vaccaro2024combinations}. To truly understand how agents as workers may reshape today's human-centered workspace, we argue for a fine-grained, workflow-level investigation spanning multiple occupations and skill domains, to enable findings that shed light on diverse areas of work.

\noindent \textbf{Inducing Computer Workflows} \quad
Several tools have been developed for recording human trajectories \citep{wang2025opencua,shaikh2025creating}. However, such raw activity data are often hard to interpret, as low-level actions are task-agnostic and lack clear semantic grounding. Even for humans themselves, articulating explicit goals for certain actions, e.g., random clicks on blank areas driven by hesitation or anxiety, can be challenging.
This poses challenges in conducting the head-to-head comparisons between human and agent computer-use activities. 
Inspired by how humans naturally communicate and coordinate around higher-level goals, we conceptualize and model activities at a coarser granularity through workflows, defined as sequences of actions frequently undertaken to achieve a specific goal \citep{wang2025agent}. Prior efforts toward this goal have largely depended on manual annotation \citep{sodhi2023heap,grunde2025designing}. In contrast, we introduce the first automated tool capable of inducing interpretable workflows from raw mouse and keyboard traces, generating structured goals at multiple levels of granularity to support diverse analytical needs.
While we primarily employ this tool to study agent and human worker behaviors in this work, it also holds promise for broader applications, such as agent skill learning and human-agent collaboration.

\section{Conclusion}
Our study reveals that today's AI agents approach human work through a distinctly programmatic lens: they translate even open-ended, visually grounded tasks into code, diverging sharply from the perceptual and interface-driven human workflows. 
This bias disrupts and slows human workflows when AI is used for automation, though its impact is far smaller in augmentation scenarios.
Despite systematic quality issues such as data fabrication and tool misuse, agent work nonetheless shows great efficiency potential that scales across domains. 
Maximizing this potential requires integrating human and agent steps based on a deeper understanding of their complementary strengths.

Looking ahead, we advocate for workflow-inspired agent development that extends programmatic reasoning beyond engineering domains, while strengthening agents’ visual grounding, workflow memory, and action calibration.
Anchoring future agent design in the study of human workflows can foster systems that operate with both rigor and understanding across the evolving landscape of human work.

\section*{Limitations}
\noindent \textbf{Coverage of Work Activities} \quad
In this study, we designed tasks to encompass a representative range of occupations and core work-related skills. We conducted significance tests to ensure the statistical reliability of all quantitative findings. 
Nonetheless, our study does not yet capture the full diversity of real-world working contexts, and thus would benefit from additional task instances with greater breadth. Meanwhile, collecting more activities from more diverse human and agent workers would help strengthen and extrapolate the findings from our study. 
As the first direct comparison of human and agent workflows, considerable effort was devoted to building the analysis toolkit and configuring the investigation dimensions, which somewhat restricted the number of tasks and work activities we could examine in depth. We view this work as an initial step toward systematic human-agent comparison, and encourage future research to expand upon it with more diverse and comprehensive task and worker coverage.

\noindent \textbf{The O*NET Database} \quad
The O*NET database provides a holistic view of occupational tasks across the U.S. labor market. However, the inherent difficulty in constructing and maintaining such a large-scale database may introduce inaccuracies. Although we utilized data from the most recent 2024 release, it may not fully reflect the current workforce distribution. Task requirements for certain occupations may have evolved, while others may have become obsolete or newly emerged in the current job market \citep{acemoglu2022artificial}.

\noindent \textbf{Understanding AI's Impact on Work} \quad
It is natural for many to wonder \textit{``will AI take my job?''} especially amid the rapid advances in AI over the past few years. Beyond offering insights to researchers of AI agents and human-AI interactions, we hope this work also speaks to the broader community of human workers whose occupations may be affected by AI adoption. Because many in the general public may have little exposure to how AI actually functions, we aim for this study to serve as an accessible and detailed account that helps demystify their capabilities and limitations --- encouraging informed awareness rather than ungrounded fear or hype.
While parts of our study may reveal the uncomfortable reality that agents may gradually take over fixed-procedure tasks as they advance, many essential aspects of human work, such as communication and decision-making, still demand substantial technological progress and regulatory considerations before comparable automation becomes feasible \citep{comunale2024Economic}. Overall, our goal is to offer a clearer, evidence-based understanding of how these seemingly powerful AI agents operate and to inform a grounded, proactive perspective on how humans and AI can move forward together.

\section*{Acknowledgments}
We would like to thank Sherry Tongshuang Wu, Yuxuan Li, Christina Qianou Ma, Zhoujun Cheng, Xuhui Zhou, Yuxiao Qu, Lawrence Jang, Xinran Zhao, Saujas Vaduguru, Vijay Viswanathan, Chenyang Yang, Mingqian Zheng, Youssouf Kebe, John Yang, Vishakh Padmakumar, Jiaxin Pei, and Alexander Spangher for their helpful discussions and insightful feedback on the project.
Zora Zhiruo Wang is supported by Google PhD Fellowship. This research is supported in part by grants from Sloan Foundation,  ONR grant N000142412532 and Open Philanthropy.

\newpage
\bibliography{main}

\newpage
\appendix

\section{Task Creation and Data Collection}
\label{app:a:data}

\subsection{Skill Annotation and Selection}
\label{app:a.1:skill-anno}

We categorize occupations and their task requirements through the scalable lens of skills. We introduce the skill annotation details by the full list of skills with exemplar tasks, and justify the selection of execution-style skills for our study.

\begin{table*}[ht]
\centering
\small
\vspace{-1mm}
\resizebox{0.99\textwidth}{!}{
\begin{tabular}{lll}
\toprule
\multicolumn{1}{c}{\bf Skill} & \multicolumn{1}{c}{\bf Occupation} & \multicolumn{1}{c}{\bf Task Requirement} \\
\midrule
\multirow{4}{*}{Data Analysis} & {Human Resources Managers} & \makecell[l]{Analyze statistical data and reports to identify and \\determine causes of personnel problems.} \\
\cmidrule{2-3}
{} & {Statisticians} & \makecell[l]{Report results of statistical analyses, including \\information in the form of graphs, charts, and tables.} \\
\midrule
\multirow{4}{*}{Engineering} & {Computer Programmers} & \makecell[l]{Write, update, and maintain computer programs or software \\packages to handle specific jobs such as tracking inventory, \\storing or retrieving data, or controlling other equipment.} \\
\cmidrule{2-3}
{} & {Web Developers} & \makecell[l]{Write supporting code for Web applications or Web sites.} \\
\midrule
\multirow{4}{*}{\makecell[l]{Administrative \\ \& Computation}} & {Bookkeeping, Accounting} & \makecell[l]{Calculate and prepare checks for utilities, taxes, and other \\payments.} \\
\cmidrule{2-3}
{} & {Legal Secretaries} & \makecell[l]{Prepare and distribute invoices to bill clients or pay account \\expenses.} \\
\midrule
\multirow{2}{*}{Writing} & {Financial Managers} & \makecell[l]{Prepare operational or risk reports for management analysis.} \\
\cmidrule{2-3}
{} & {Judicial Law Clerks} & \makecell[l]{Draft or proofread judicial opinions, decisions, or citations.} \\
\midrule
\multirow{4}{*}{Design} & {Graphical Designers} & \makecell[l]{Draw and print charts, graphs, illustrations, and other \\artwork, using computer.} \\
\cmidrule{2-3}
{} & {Architects} & \makecell[l]{Prepare scale drawings or architectural designs, \\using computer-aided design or other tools.} \\
\midrule
\multirow{4}{*}{Communication} & \makecell[l]{Advertising and Promotions \\Managers} & {Coordinate with the media to disseminate advertising.} \\
\cmidrule{2-3}
{} & {Health Informatics Specialists} & \makecell[l]{Translate nursing practice information between nurses and \\systems engineers, analysts, or designers, using object-\\oriented models or other techniques.} \\
\midrule
\multirow{3}{*}{Decision-Making} & \makecell[l]{Computer and Information \\Research Scientists} & {Approve, prepare, monitor, and adjust operational budgets.} \\
\cmidrule{2-3}
{} & {Sustainability Specialists} & \makecell[l]{Review and revise sustainability proposals or policies.} \\
\bottomrule
\end{tabular}
}
\caption{Skill categories with exemplar occupations and tasks from O*NET.}
\label{tab:skill-annotation}
\end{table*}

As shown from the exemplar occupations and task requirements in \autoref{tab:skill-annotation}, the top five skills are of \textit{execution} style that aim to finish some work with deliverables, while the bottom ones are more of \textit{coordination} style that organize between workers. Since our study mostly focuses on studying single, autonomous agent workers, we do not consider the communication and decision-making tasks. Despite filtering these skills out, our final task pool is representative of the 287 computer-related occupations and 71.9\% of their daily tasks.

\subsection{Task Instantiation}
\label{app:a.2:task-instantiation}

While TAC tasks suffice for some fixed-procedure skill domains, we need to synthesize two additional tasks to ensure coverage for more open-ended skill categories, namely writing (\texttt{finance-prepare-quarterly-report}) and design (\texttt{design-company-landing-page}).

Refer to the full list of tasks in \autoref{tab:task-instantiation}.

\begin{table*}[ht]
\centering
\small
\vspace{-1mm}
\resizebox{0.99\textwidth}{!}{
\begin{tabular}{ll}
\toprule
\multicolumn{1}{c}{\bf Skill} & \multicolumn{1}{c}{\bf Task (Abbreviated)} \\
\midrule
{Data Analysis} & \makecell[l]{\href{https://github.com/TheAgentCompany/TheAgentCompany/tree/main/workspaces/tasks/ds-predictive-modeling}{ds-predictive-modeling}, \href{https://github.com/TheAgentCompany/TheAgentCompany/tree/main/workspaces/tasks/ds-stock-analysis-slides}{ds-stock-analysis-slides}, \\\href{https://github.com/TheAgentCompany/TheAgentCompany/tree/main/workspaces/tasks/finance-budget-variance}{finance-budget-variance}, \href{https://github.com/TheAgentCompany/TheAgentCompany/tree/main/workspaces/tasks/hr-check-attendance-multiple-days-department}{hr-check-attendance-multiple-days}} \\
\midrule
{Engineering} & \makecell[l]{\href{https://github.com/TheAgentCompany/TheAgentCompany/tree/main/workspaces/tasks/sde-report-unit-test-coverage-to-plane}{sde-report-unit-test-coverage}, \href{https://github.com/TheAgentCompany/TheAgentCompany/tree/main/workspaces/tasks/sde-run-janusgraph}{sde-run-janusgraph}, \\ \href{https://github.com/TheAgentCompany/TheAgentCompany/tree/main/workspaces/tasks/sde-troubleshoot-dev-setup}{sde-troubleshoot-dev-setup}, \href{https://github.com/TheAgentCompany/TheAgentCompany/tree/main/workspaces/tasks/sde-update-dev-document}{sde-update-dev-document}} \\
\midrule
{Computation \& Administrative} & \makecell[l]{\href{https://github.com/TheAgentCompany/TheAgentCompany/tree/main/workspaces/tasks/finance-create-10k-income-report}{finance-create-10k-income-report}, \href{https://github.com/TheAgentCompany/TheAgentCompany/tree/main/workspaces/tasks/finance-expense-validation}{finance-expense-validation}, \\\href{https://github.com/TheAgentCompany/TheAgentCompany/tree/main/workspaces/tasks/finance-invoice-matching}{finance-invoice-matching}, \href{https://github.com/TheAgentCompany/TheAgentCompany/tree/main/workspaces/tasks/hr-analyze-outing-bills}{hr-analyze-outing-bills}} \\
\midrule
{Writing} & {finance-prepare-quarterly-report, \href{https://github.com/TheAgentCompany/TheAgentCompany/tree/main/workspaces/tasks/hr-new-grad-job-description}{hr-new-grad-job-description}} \\
\midrule
{Design} & {design-company-landing-page, \href{https://github.com/TheAgentCompany/TheAgentCompany/tree/main/workspaces/tasks/sde-create-new-gitlab-project-logo}{sde-create-new-gitlab-project-logo}} \\
\bottomrule
\end{tabular}
}
\caption{Specific tasks instantiated for each skill category.}
\label{tab:task-instantiation}
\end{table*}

\subsection{Human Activity Collection}
\label{app:a.3:human-activity}

\noindent \textbf{Human Worker Recruitment} \quad
We hire human workers who have relevant educational backgrounds and are currently doing jobs related to each task and skill we study.
We enforce this qualification screening as it would be more informative to compare agents against proficient human workers, for the purpose of studying how far agents are in automating human work.
We hire human workers from Upwork, and perform initial screening of human workers based on their professional qualifications, without accessing any other personal information. 
We hire 3 human workers for each task. Human workers can use any software or tools they prefer to complete their work, including professional software and AI tools, to ensure that they are working in their most productive scenario and can serve as good canonicals for agent workers.
We do not explicitly control the expertise levels of human workers, as it can be somewhat hard to measure without a long-term interaction with the worker. Nonetheless, we do observe varied expertise among human workers, as reflected in their work efficiency and quality (\S\ref{sec:5:analysis-quality}).

\noindent \textbf{Processing Raw Computer-Use Activities} \quad
Computer-use activities collected from human real-world usage can be redundant, due to the limitations in the recording tool implementation. 
For example, each mouse scrolling (\texttt{scroll()}) and keyboard press (\texttt{keypress(`h')}) event is recorded as an independent action. These human actions execute with arguments that are more fine-grained (e.g., \texttt{keypress(`h')}, \texttt{keypress(`o')}, $\cdots$, \texttt{keypress(`?')}) than arguments in the agent actions (e.g., \texttt{keypress(`how ai agents do human work?')}), thus becoming costly to process and hard to interpret. 
We hence perform post-processing to merge consecutive \texttt{keypress} and \texttt{scroll} actions into single actions with concatenated arguments. We also identify possible double clicks if two \texttt{click()} actions happen within 0.1 seconds; and merge them into a single \texttt{double\_click()} action. 
This effectively simplifies the trajectory data, reducing the action count by 83.2\% (from 5831.1 to 981.1 actions) in an average human trajectory, meanwhile aligning better with the granularity of agent computer activities \citep{wang2025opencua}.

\subsection{Agent Activity Collection}
\label{app:a.4:agent-activity}
While we can directly access and store all actions and states for OpenHands agents, we can only access the information presented at the ChatGPT and Manus agent UI interfaces. For each step in the ChatGPT and Manus trajectories, the agent will elaborate its thought process in text, and present a screenshot of the virtual computer screen. We infer the actions taken from the agent's thoughts by mapping back to the common computer-use agent action space such as \citet{zhou2024webarena,xie2024osworld}. We present the actions employed by all agent frameworks in \autoref{tab:agent-action-space}. Nonetheless, action names for ChatGPT and Manus may not be precisely what was executed by the agent, but are our best guesses and should serve a similar functionality to the actual action, which we unfortunately do not have access to. Meanwhile, the listed actions are only actions that we observe agents have taken while completing all the task instances in our study. These actions may not constitute the full action space available to the agents.

Also, for closed-source agent frameworks (ChatGPT and Manus), since the detailed timestamp associated with each step is not provided, we can only obtain time estimations to the precision of minutes, so there may be an error of up to 60 seconds in the recorded time.

\begin{table*}[t!]
\centering
\small
\vspace{-1mm}
\resizebox{0.99\textwidth}{!}{
\begin{tabular}{llc}
\toprule
\multicolumn{1}{c}{\bf Action} & \multicolumn{1}{c}{\bf Description} & {\bf Agent} \\
\midrule
\texttt{browse\_interactive} & Navigate and interact with web pages in a browser. & {\oh \oai \manus} \\
\texttt{click} & {Click on an element on the computer screen.} & {\oh \oai \manus} \\
\texttt{right\_click} & {Right click on an element on the computer screen.} & {\oh \oai \manus} \\
\texttt{double\_click} & {Double-click on an element on the computer screen.} & {\oh \oai \manus} \\
\texttt{noop} & {Do not take actions for the current iteration.} & {\oai \manus} \\
\texttt{scroll} & {Scroll up or down on a webpage in browser.} & {\oh \oai \manus} \\
\texttt{zoom} & {Zoom in or zoom out of a page.} & {\oai} \\
\texttt{keypress} & {Type in text to an element on the computer screen.} & {\oh \oai \manus} \\
\texttt{run\_ipython} & Execute a Python code snippet in Jupyter Notebook. & {\oh} \\
\texttt{search} & Perform search online to gather information. & {\oh \oai \manus} \\
\texttt{read} & Read the content of files in the terminal. & {\oh \manus} \\
\texttt{create} & Create a new file with specified content in the terminal. & {\manus} \\
\texttt{edit} & Edit file content by executing bash command in terminal. & {\oh \oai \manus} \\
\texttt{run} & Execute bash command in the terminal. & {\oh \oai \manus} \\
\texttt{think} & Perform thinking and deliver a chain of thought to user. & {\oh \oai \manus} \\
\texttt{message} & Send a text message to user. & {\oh \oai \manus} \\
\texttt{task\_tracking} & Track the task-solving progress by managing a todo list. & {\oh} \\
\texttt{open\_image} & {Open and view an image file.} & {\oai \manus} \\
\texttt{search\_image} & {Search for images online.} & {\manus} \\
\texttt{generate\_image} & {Generate an image.} & {\oai \manus} \\
\texttt{reset\_environment} & {Reset the virtual computer environment.} & {\manus} \\
\bottomrule
\end{tabular}
}
\caption{Action space of OpenHands \oh, ChatGPT \oai, and Manus \manus $~~$ agents.}
\label{tab:agent-action-space}
\end{table*}

\section{Workflow Induction and Verification}
\label{app:b:workflow-prompts}

We provide the detailed prompts used throughout every LM-involved step during workflow induction and verification. We use \texttt{claude-sonnet-3.7} for all steps detailed below.

\subsection{Workflow Induction}
\label{app:b.1:induction-prompt}
We prompt model to merge adjacent activity segments (\S\ref{sec:3.2:workflow-induction}) using the following prompt:

\begin{tcolorbox}[colback=gray!10, colframe=gray!60, coltitle=black, title=Prompt for Merging Semantically Coherent Segments, fonttitle=\bfseries]
Your task is to determine if the two computer screens focus on the same software.
For each screen, first identify the software it is focused on in the front, e.g., Google Chrome, VSCode, Finder, etc.
Then, compare the software on the two screens. If they are the same, output `YES'. Otherwise, output `NO'.
\end{tcolorbox}

\subsection{Workflow Quality Evaluation}
\label{app:b.2:eval-prompt}

For each workflow step, we measure the following two metrics:

\noindent \textbf{Action-Goal Consistency} \quad
We provide the evaluator with the (i) subgoal uttered in NL and (ii) action sequence along with annotated intentions; and ask the evaluator to output if any actions are irrelevant to the main subgoal, in a binary \texttt{YES}/\texttt{NO} format.
For large-scale automatic evaluation, we use \texttt{claude-sonnet-3.7} as the evaluator with the following prompt:

\begin{tcolorbox}[colback=gray!10, colframe=gray!60, coltitle=black, title=Prompt for Evaluating Goal-Action Consistency, fonttitle=\bfseries]
Your task is to determine if the action sequence aligns with the task goal. The actions may not have achieved the goal yet, return 'YES' as long as it attempts to progress towards the goal. If no action sequence is provided, return 'YES'.
\end{tcolorbox}

\noindent \textbf{Modularity} \quad
We use \texttt{claude-sonnet-3.7} as the evaluator with the following prompt:

\begin{tcolorbox}[colback=gray!10, colframe=gray!60, coltitle=black, title=Prompt for Evaluating Modularity, fonttitle=\bfseries]
Evaluate whether the current task-solving step is clearly focused on causally consistent procedures or achieving the same final goal, instead of a concatenation of multiple distinct topics or interfaces.
\end{tcolorbox}

\section{Workflow Analyses}
\label{app:c:workflow-analysis}

\subsection{Do Agents Follow Human Task Workflows?}
\label{app:c.1:workflow-alignment}

\begin{wraptable}[11]{r}{0.60\textwidth}
\centering
\small
\vspace{-2mm}
\resizebox{0.60\textwidth}{!}{
\begin{tabular}{l|cccc}
\toprule
{\bf Metric} & \textbf{\makecell{Agent - \\ Agent}} & \textbf{\makecell{Human - \\ Human}} & \textbf{\makecell{Agent - \\ Human}} & \textbf{\makecell{Capable Agent - \\ Independent Human}} \\
\midrule
{match\%} & {70.2} & {83.5} & {74.6} & {84.4} \\
\midrule
{t-stats} & {4.592} & {8.111} & {8.879} & {17.794} \\
{p value} & {0.000} & {0.000} & {0.000} & {0.000} \\
\bottomrule
\end{tabular}
}
\vspace{-1mm}
\caption{Workflow alignment between workers. Capable agent means those who can successfully finish the task till the end. Independent human stands for those who finish the task by themselves without using any AI-related tools.}
\label{tab:workflow-alignment-worker-groups}
\vspace{2mm}
\end{wraptable}

\paragraph{Human and Agent Workflows Largely Align}
\autoref{tab:workflow-alignment-worker-groups} presents the workflow alignment measures among agents, among humans, and between human and agent workers.
Human and agent workflows share 83.0\% of steps, with step orders preserved for 99.8\% of the time --- both indicating a high degree of alignment between.

\noindent \textbf{Capable Agents Match Independent Human Workflows} \quad
Reading from the rightmost column in \autoref{tab:workflow-alignment-worker-groups}, the workflows of capable agents (i.e., agents that can finish the task till the end without getting stuck in the middle) and independent humans (i.e., humans that conduct the task without AI automation) exhibit a significantly higher workflow alignment of 83.4\%, compared to the alignment of all human and agent workers (74.5\%), persistently across all skill categories.

\begin{wraptable}[8]{r}{0.60\textwidth}
\centering
\small
\vspace{-2mm}
\resizebox{0.60\textwidth}{!}{
\begin{tabular}{l|ccccc}
\toprule
{\bf Metric} & {\bf DS} & {\bf Eng} & {\bf Comp} & {\bf Writing} & {\bf Design} \\
\midrule
{match\%} & {84.5} & {90.0} & {88.0} & {80.3} & {72.1} \\
\midrule
{t-stats} & {12.359} & {11.247} & {10.837} & {7.974} & \markcell{1.051} \\
{p value} & {0.001} & {0.001} & {0.001} & {0.040} & \markcell{0.242} \\
\bottomrule
\end{tabular}
}
\vspace{-1mm}
\caption{Task breakdown of agent workflow alignment.}
\label{tab:workflow-alignment-task-breakdown}
\vspace{2mm}
\end{wraptable}

\noindent \textbf{Open-Ended Tasks Drive More Divergent Workflows} \quad
While we have shown that human and agent workflows largely align in \S\ref{sec:4:analysis-workflow}, this alignment is especially prominent for more fixed-procedure tasks, including data analysis, engineering, and computation, where the human-agent alignment scores are significantly above chance (50\% alignment).
As the tasks become more open-ended in nature (e.g., $\rightarrow$ writing $\rightarrow$ design), human-agent alignment is less substantial, signaling increasing divergence between human and agent behaviors in solving these tasks.
Particularly for design tasks where significant workflow alignment is no longer observed, we find agents mostly focusing on program-related steps such as `setup Figma environment' and `create responsive HTML page prototype', while humans focus on ideation and visuals such as `browse community templates' and `adjust elements on the canvas'.


\begin{wrapfigure}[14]{r}{0.5\textwidth}
\vspace{-3mm}    
\includegraphics[width=0.5\textwidth]{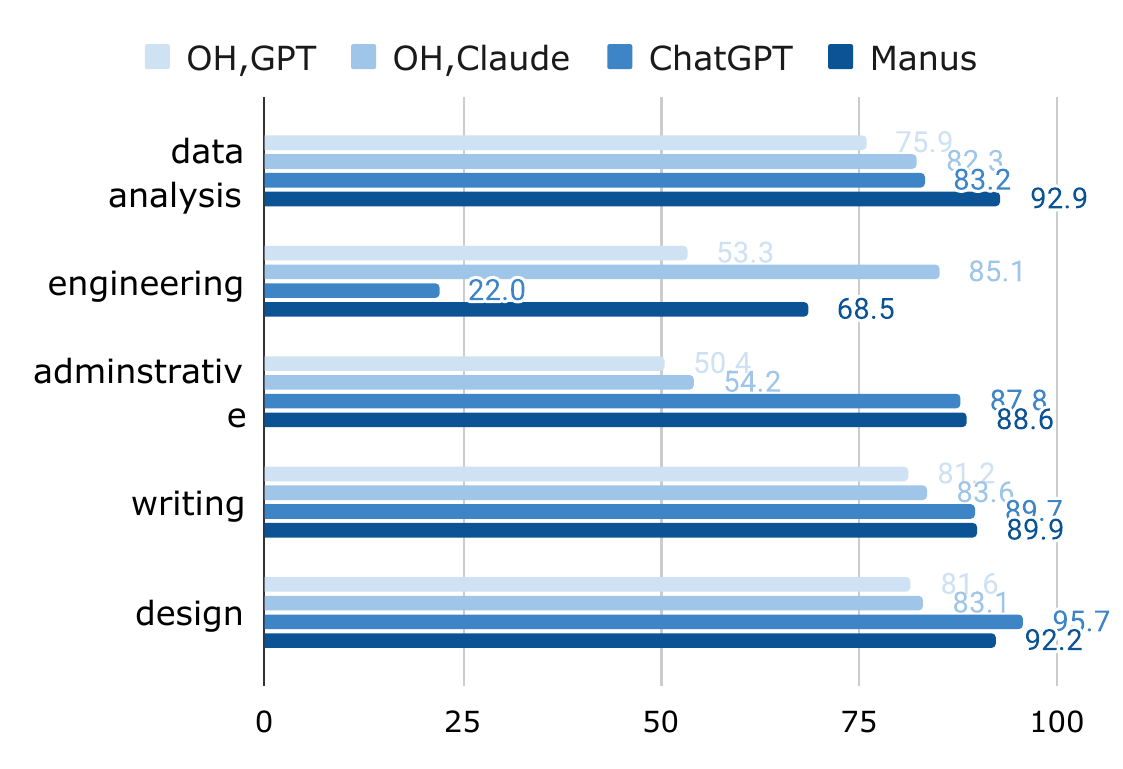}
\vspace{-5mm}
\caption{Progress of different agents on varied work-related tasks. OH stands for OpenHands.}
\label{fig:agent-progress}
\end{wrapfigure}

\noindent \textbf{Majorly Differ in Workflow Progress} \quad
We evaluate how far agent workflows go with respect to human workflows, if grounding human activities as reference solutions. Specifically, comparing an agent workflow against an individual human workflow, we identify the farthest agent step that matches a human step, and calculate the percentage of actions till that point within the agent's trajectory. We denote this as the agent's \textit{progress} with respect to a `canonical' human trajectory.

As shown in \autoref{fig:agent-progress}, the Manus agent makes the most progress, finishing 86.4\% of the tasks across skill categories. 
The open-source OpenHands agents make less progress in general, 68.5\% and 77.7\% with \texttt{gpt-4o} and \texttt{claude-sonnet-4} LM backbones. Nonetheless, on engineering tasks, it achieves better performance than ChatGPT and Manus when using the same LM backbone, demonstrating its advantages as a coding-oriented agent framework.
Corroborated by our manual inspection, this variance in agent progress largely explains the greater misalignment among agents (70.2\%) than among humans (83.5\%) in \autoref{tab:workflow-alignment-worker-groups}.


\begin{wrapfigure}[10]{r}{0.5\textwidth}
\vspace{-4mm}    
\includegraphics[width=0.5\textwidth]{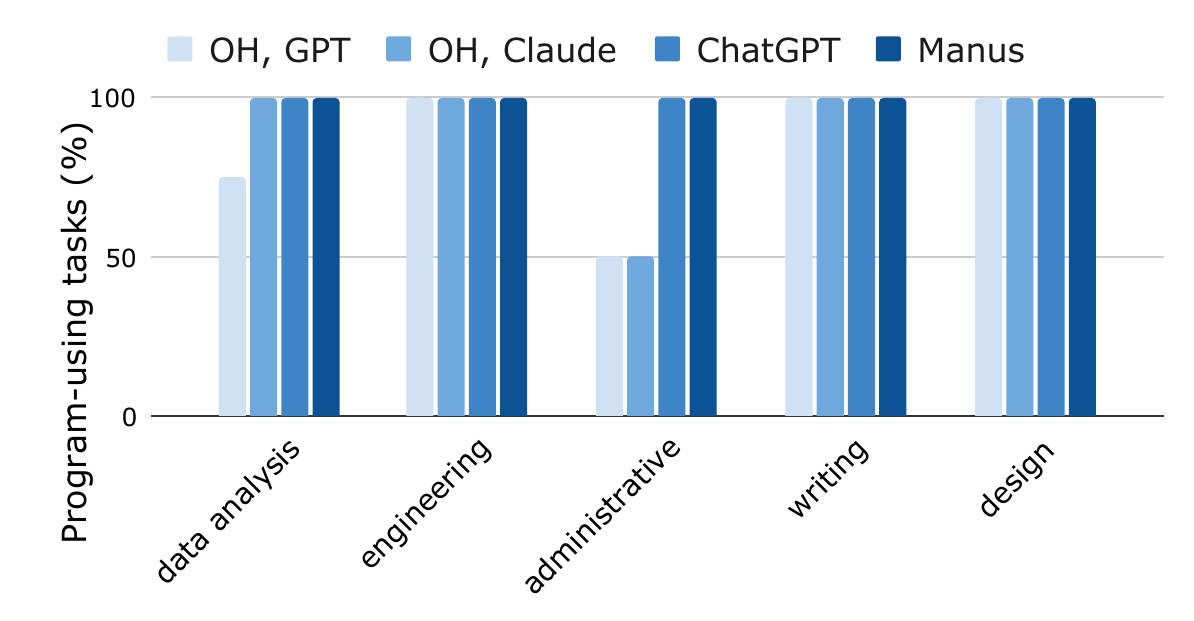}
\vspace{-8mm}
\caption{Agents program to solve all tasks.}
\label{fig:agent-program-use-rate}
\end{wrapfigure}

\begin{figure*}[t!]
\centering
    \includegraphics[width=0.95\textwidth]{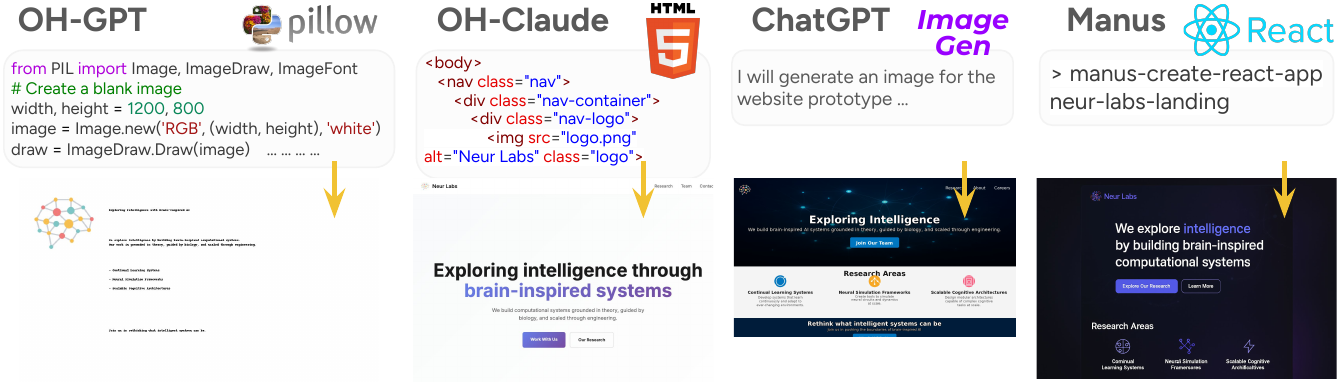}
\caption{Agent workers use diverse programming tools.}
\label{fig:tool-vary}
\end{figure*}

\noindent \textbf{Agents Are Programmatic Workers} \quad
In contrast to human workers, agents seldom execute tasks via the UI interface. Instead, agents always employ alternative programmatic tools.
\autoref{fig:tool-vary} (right) illustrates agents' program use rate across different task categories. As clearly shown by the 100\% program-use rate across categories, this programmatic bias extends beyond engineering tasks to encompass virtually all computer-based activities, including those not inherently programmable, such as administrative or design work.
In fact, for all the data analysis and administrative tasks where certain OpenHands agents do not use programs, they stalled at the initial file navigation step within the file drive interface, without having the chance to solve the core analysis or processing step yet. Based on their behavior on similar tasks within the same skill domain, it is highly likely that these agents would have adopted a programmatic approach had they progressed beyond file navigation.

\noindent \textbf{Agents Align Better with Program-Using Human Workers} \quad
To concretely test this hypothesis, we measure the alignment score of agent workflows to human workflows that use programmatic tools and non-programmatic tools, respectively.
As the high-level steps match regardless of the tools employed, we analyze steps one level deeper in the workflow hierarchy (as illustrated in \autoref{fig:workflow-overall}). Specifically, given an agent-human workflow pair, for each of the matched high-level steps, we compare the alignment score for the sub-steps a level lower in the workflow hierarchy.

\begin{wraptable}[11]{r}{0.40\textwidth}
\centering
\small
\vspace{-2mm}
\resizebox{0.40\textwidth}{!}{
\begin{tabular}{l|cc}
\toprule
\multicolumn{1}{c|}{\bf Agent} & \textbf{\makecell{Human \\w/ Program}} & \textbf{\makecell{Human \\w/o Program}} \\
\midrule
{Overall} & {34.9} & {$~~$7.1} \\
\midrule
{OH, GPT} & {60.0} & {10.0} \\
{OH, Claude} & {29.6} & {$~~$5.0} \\
{ChatGPT} & {20.0} & {$~~$3.3} \\
{Manus} & {30.0} & {10.0} \\
\bottomrule
\end{tabular}
}
\vspace{-2mm}
\caption{\small Fine-grained agent workflows align better with program-using human workers than non-programmers.}
\label{tab:alignment-wrt-program}
\vspace{2mm}
\end{wraptable}

As shown in \autoref{tab:alignment-wrt-program}, agent workers exhibit stronger alignment with program-using human steps than those using non-programmatic ones. This highlights a defining characteristic of agent workers --- a programmatic mode of operation.
This is not only true for coding-oriented OpenHands agents, but also for general-purpose computer-use agents such as ChatGPT and Manus.
While this programmatic tendency is understandable in tasks like data analysis --- domains well represented in LMs trained on code data --- this programmatic behavior notably persists even in visually intensive tasks such as slide creation and graphical design.
Not only is such `blind' visual creation possible, but LM-supported agents may in fact be more proficient at editing in the symbolic space (i.e., write programs) than in the visual space (i.e., adjust pixels) \citep{qiu2024can}.

\subsection{Tool Usage}
\label{app:c.2:tool-usage}

\paragraph{Agents Use Diverse Programmatic Tools}
Despite all agents program their work all the time, this does not imply that their task-solving procedures are homogeneous. In fact, closer inspection of their tool usage reveals considerable diversity: agents employ a range of programmatic tools, many of which are custom-built by their developers. For example, in the design company landing page task shown in \autoref{fig:tool-vary}, OH-GPT uses basic \texttt{PIL.Image} drawing, OH-Claude uses \texttt{HTML}, ChatGPT uses an internal image generation tool, and Manus uses a specially-designed ReAct program. This diversity in tool usage results in webpages with varied characteristics, for instance, PIL-drawing tends to be less website-like, pixel-based image-gen tool may produce illegible text. Notably, product-oriented agents are often equipped with specialized tools tailored to specific tasks, such as website design or slide making. This suggests that developing task-oriented programming tools, as functional equivalents to human-preferred UI interfaces, could be an effective way to build more capable agent workers.

We illustrate the diverse tool usage of workers on one exemplar data analysis task in \S\ref{sec:4.2:tool-affordance}. Here we provide the tool usage visualization for all tasks involved in our study, to show the general trend of diverse tool usage across different skill categories, and may provide further insights in other directions. 

\begin{figure*}[ht]
\centering
    \includegraphics[width=0.49\textwidth]{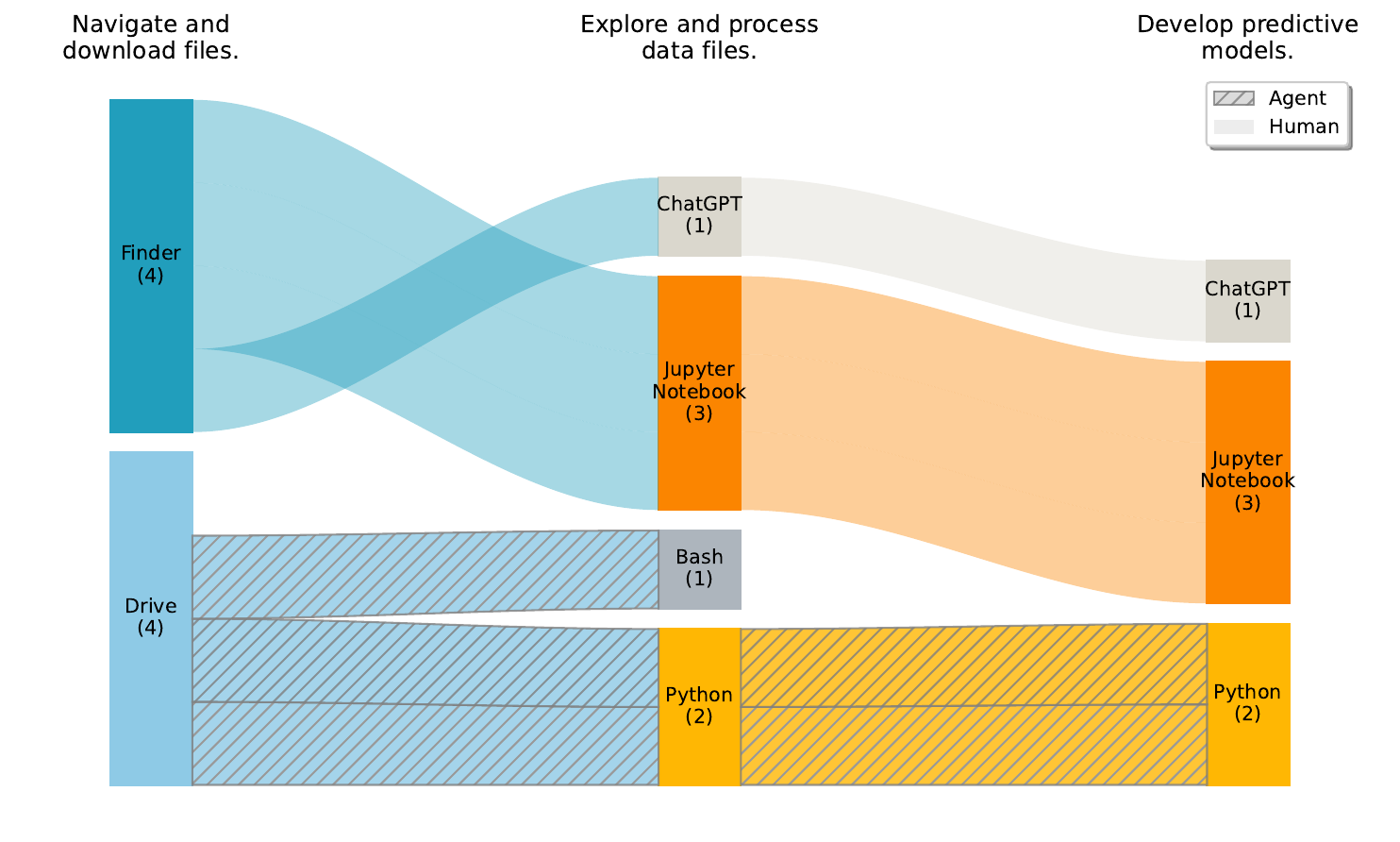}
    \hfill
    \includegraphics[width=0.49\textwidth]{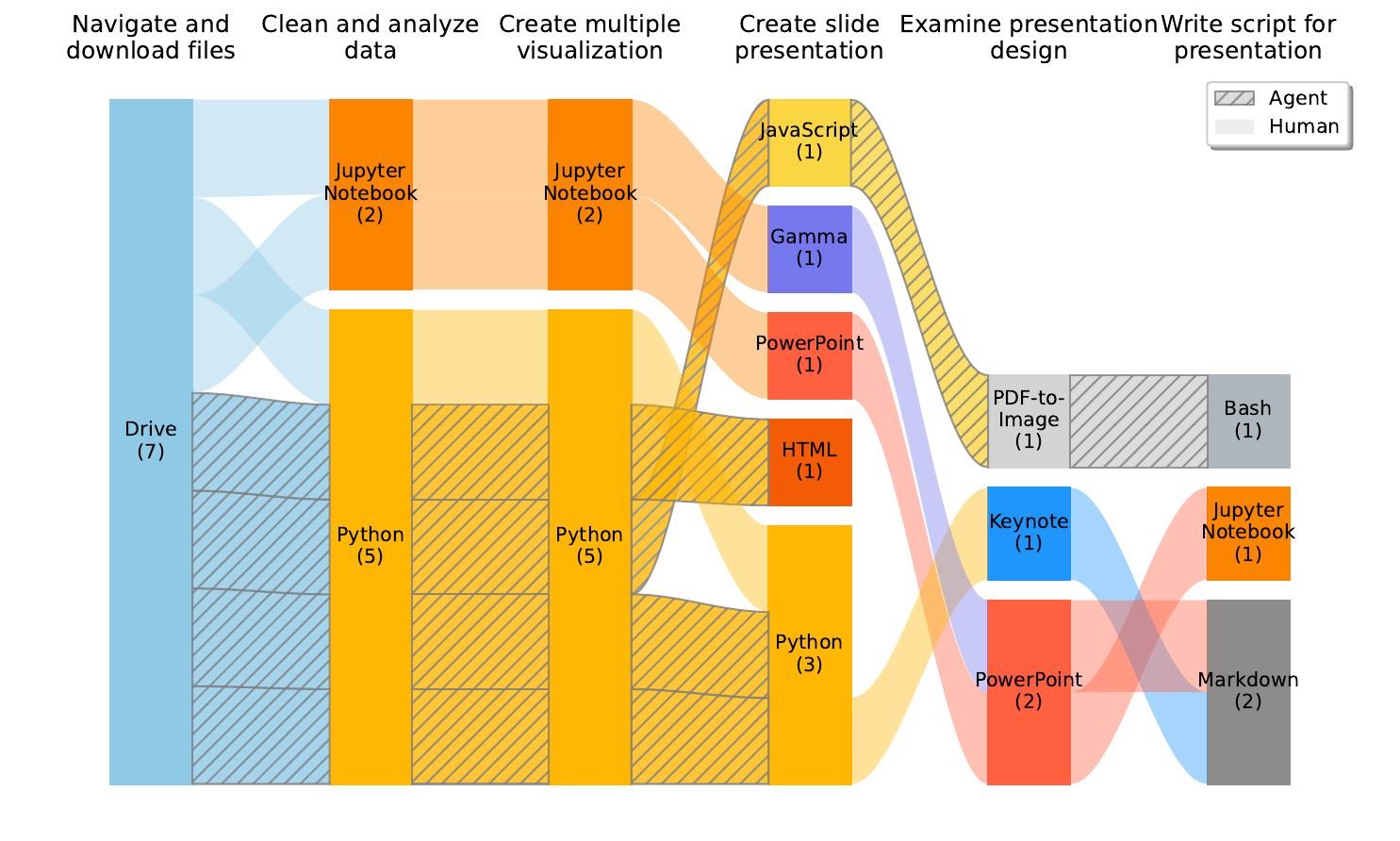}
\caption{\small{Tool usage in \texttt{ds-predictive-modeling} (left) and \texttt{ds-stock-analysis-slides} (right).}}
\label{fig:tool-use-12}
\end{figure*}

\begin{figure*}[ht]
\centering
    \includegraphics[width=0.49\textwidth]{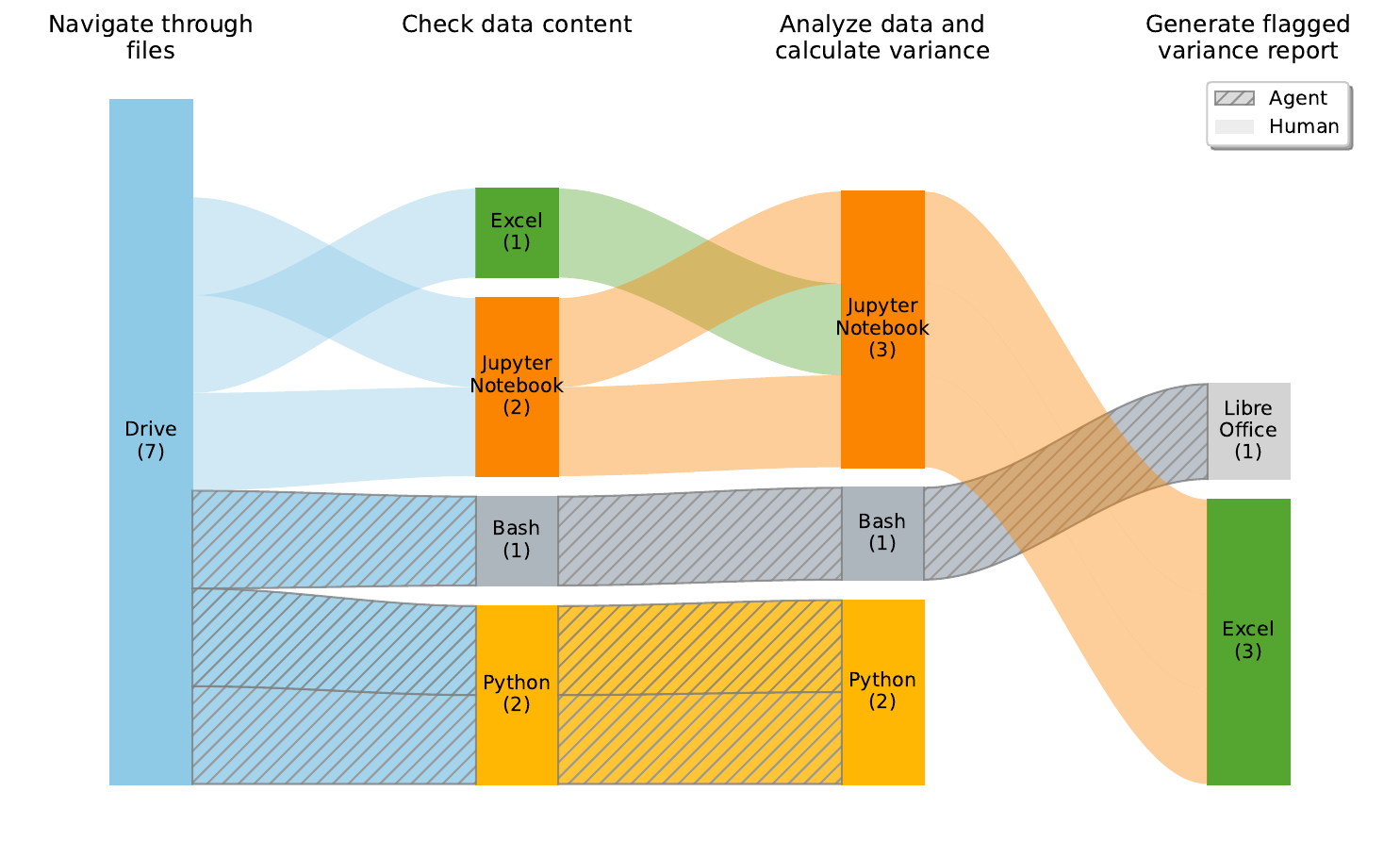}
    \hfill
    \includegraphics[width=0.49\textwidth]{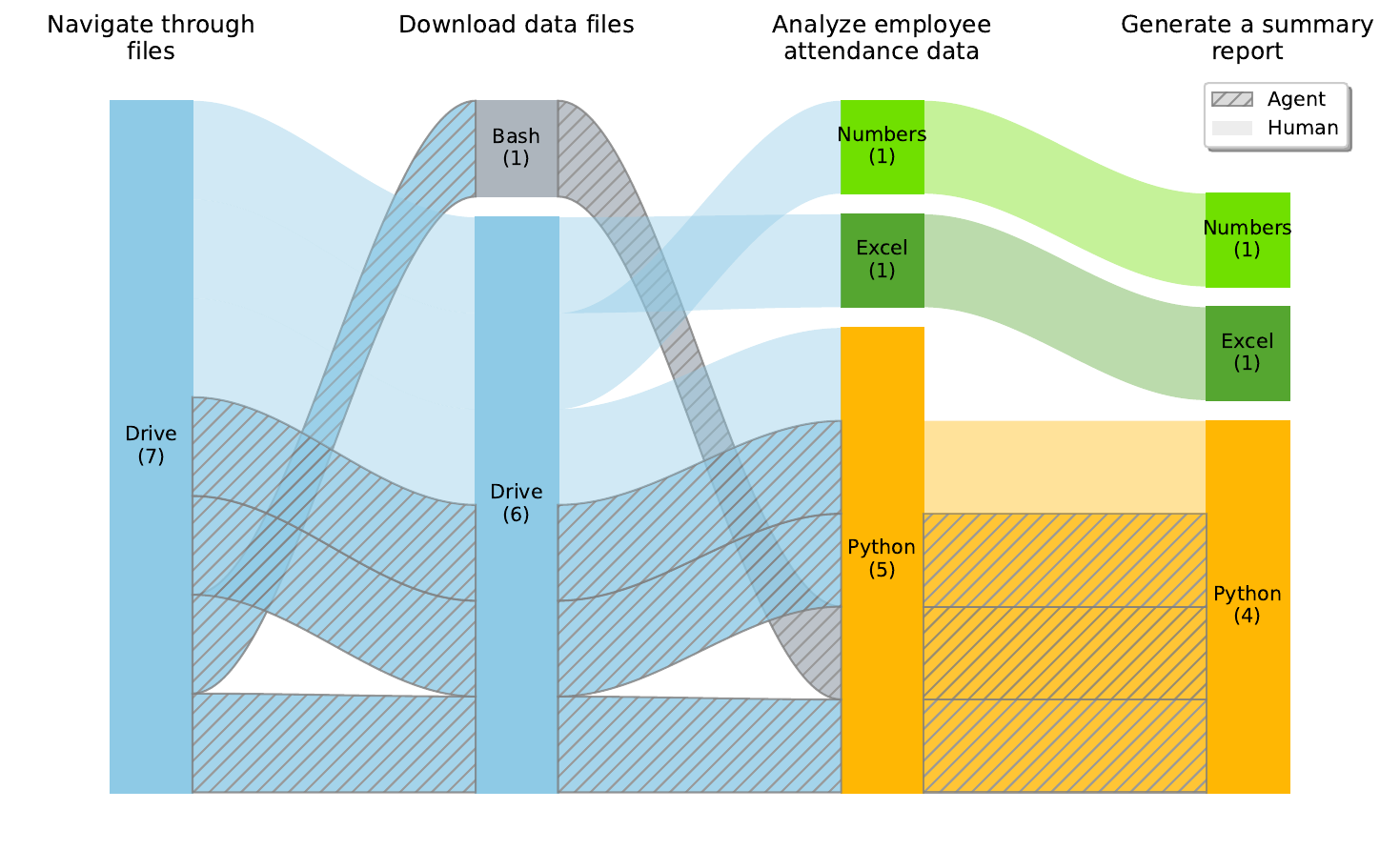}
\caption{\small{Tool usage in \texttt{finance-budget-variance} (left) and \texttt{hr-check-attendance} (right).}}
\label{fig:tool-use-34}
\end{figure*}

\begin{figure*}[ht]
\centering
    \includegraphics[width=0.49\textwidth]{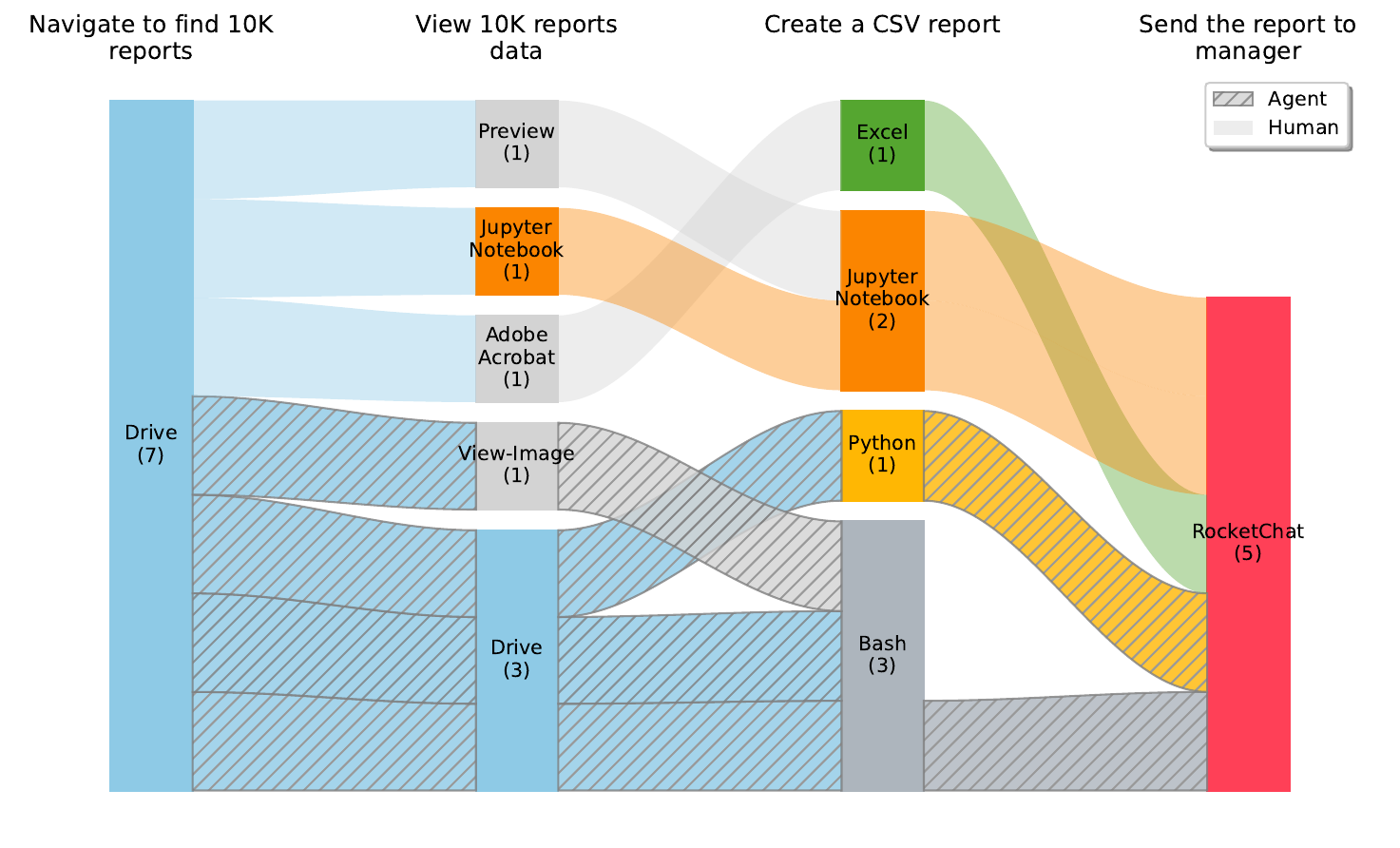}
    \hfill
    \includegraphics[width=0.49\textwidth]{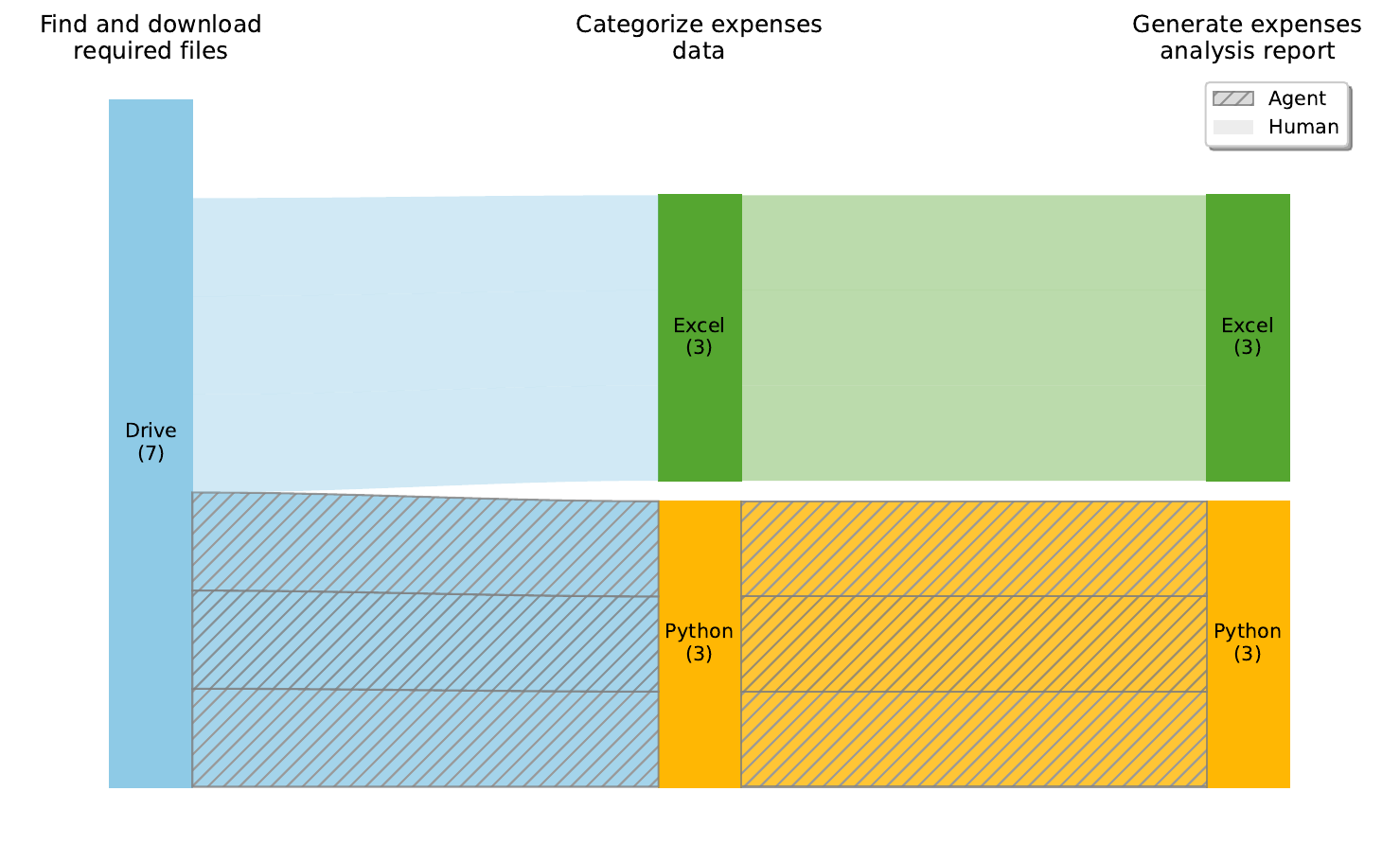}
\caption{\small{Tool usage in \texttt{create-10k-report} (left) and \texttt{finance-expense-validation} (right).}}
\label{fig:tool-use-56}
\end{figure*}

\begin{figure*}[ht]
\centering
    \includegraphics[width=0.49\textwidth]{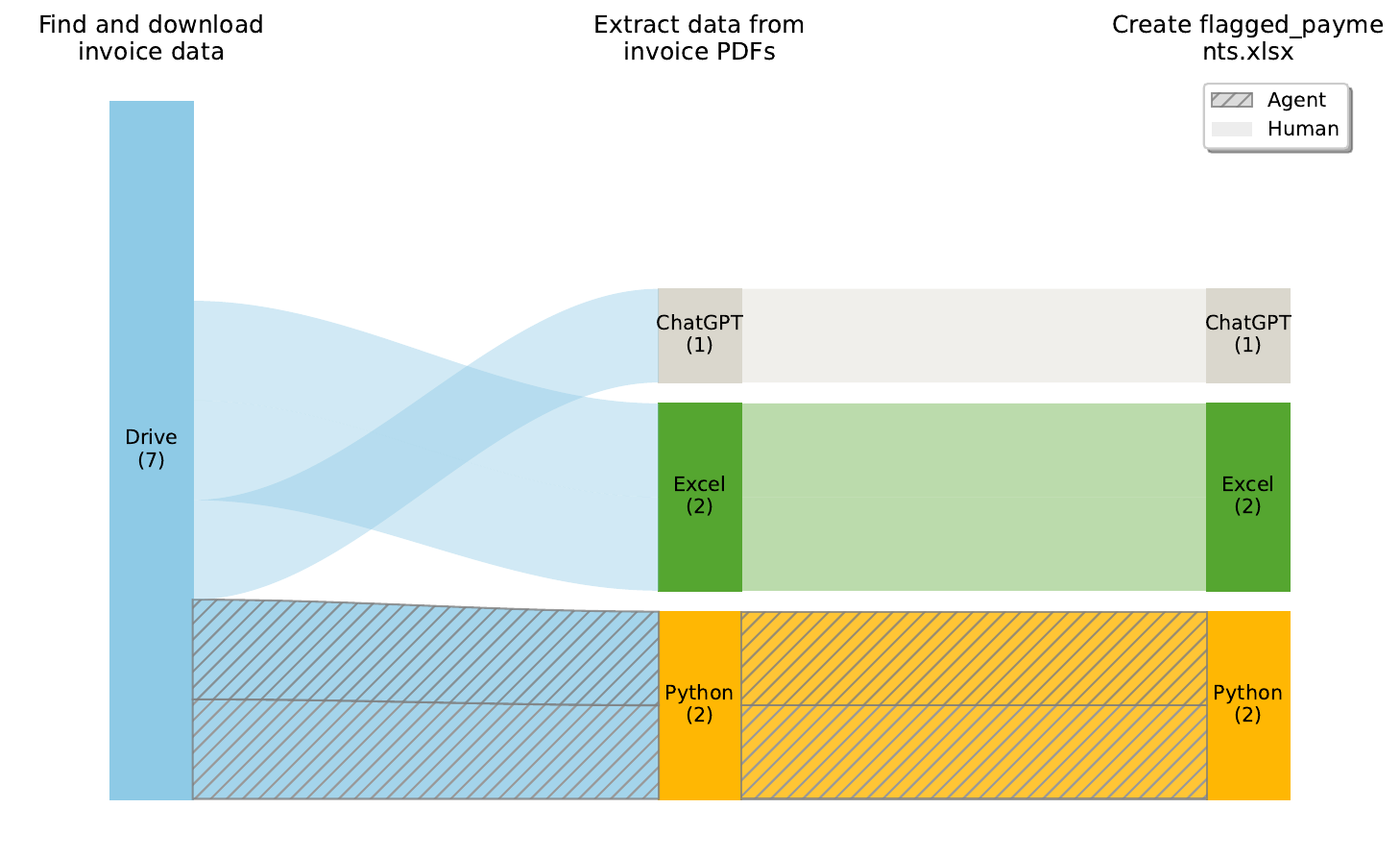}
    \hfill
    \includegraphics[width=0.49\textwidth]{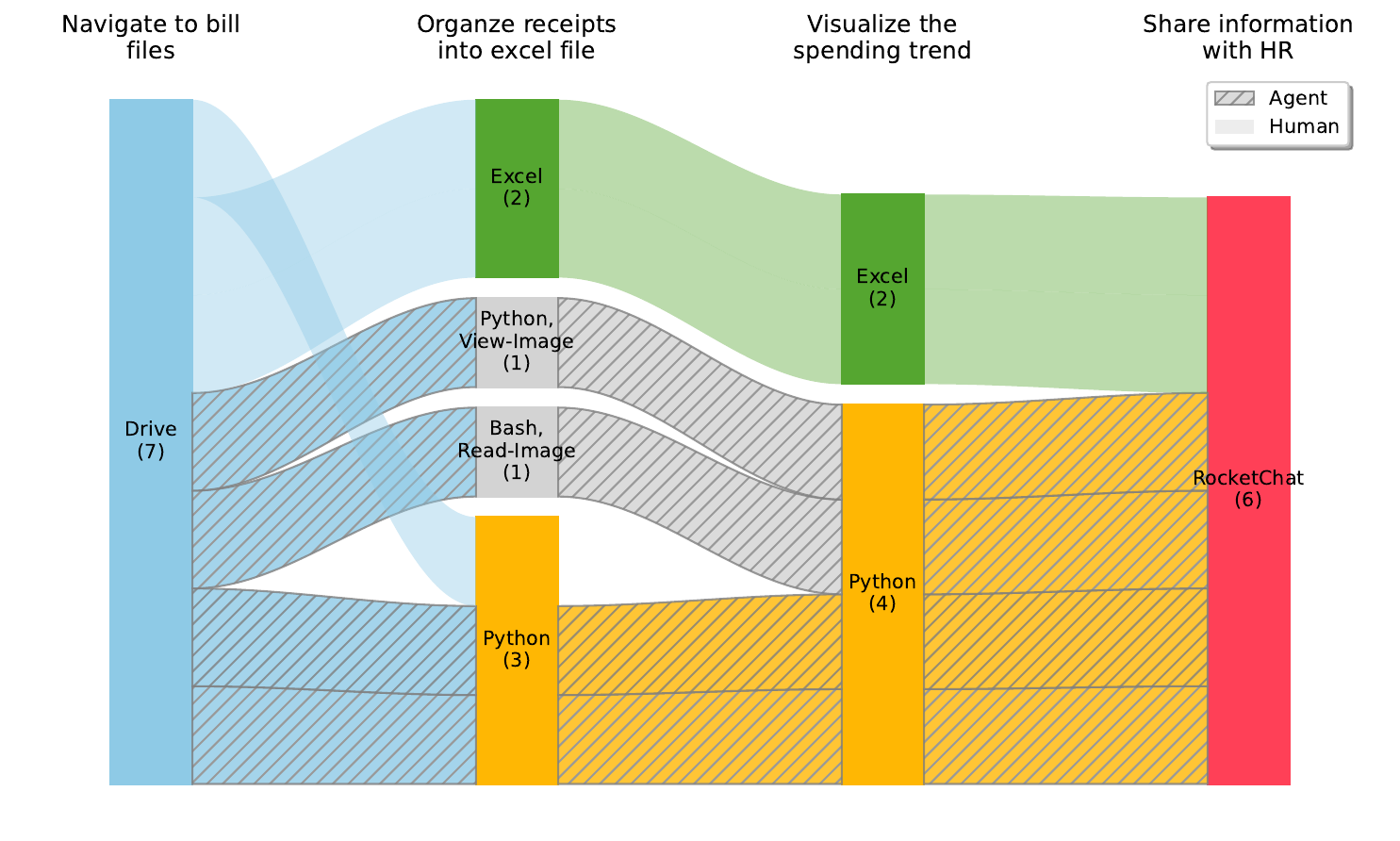}
\caption{\small{Tool usage in \texttt{finance-invoice-matching} (left) and \texttt{hy-analyze-outing-bills} (right).}}
\label{fig:tool-use-78}
\end{figure*}

\begin{figure*}[ht]
\centering
    \includegraphics[width=0.49\textwidth]{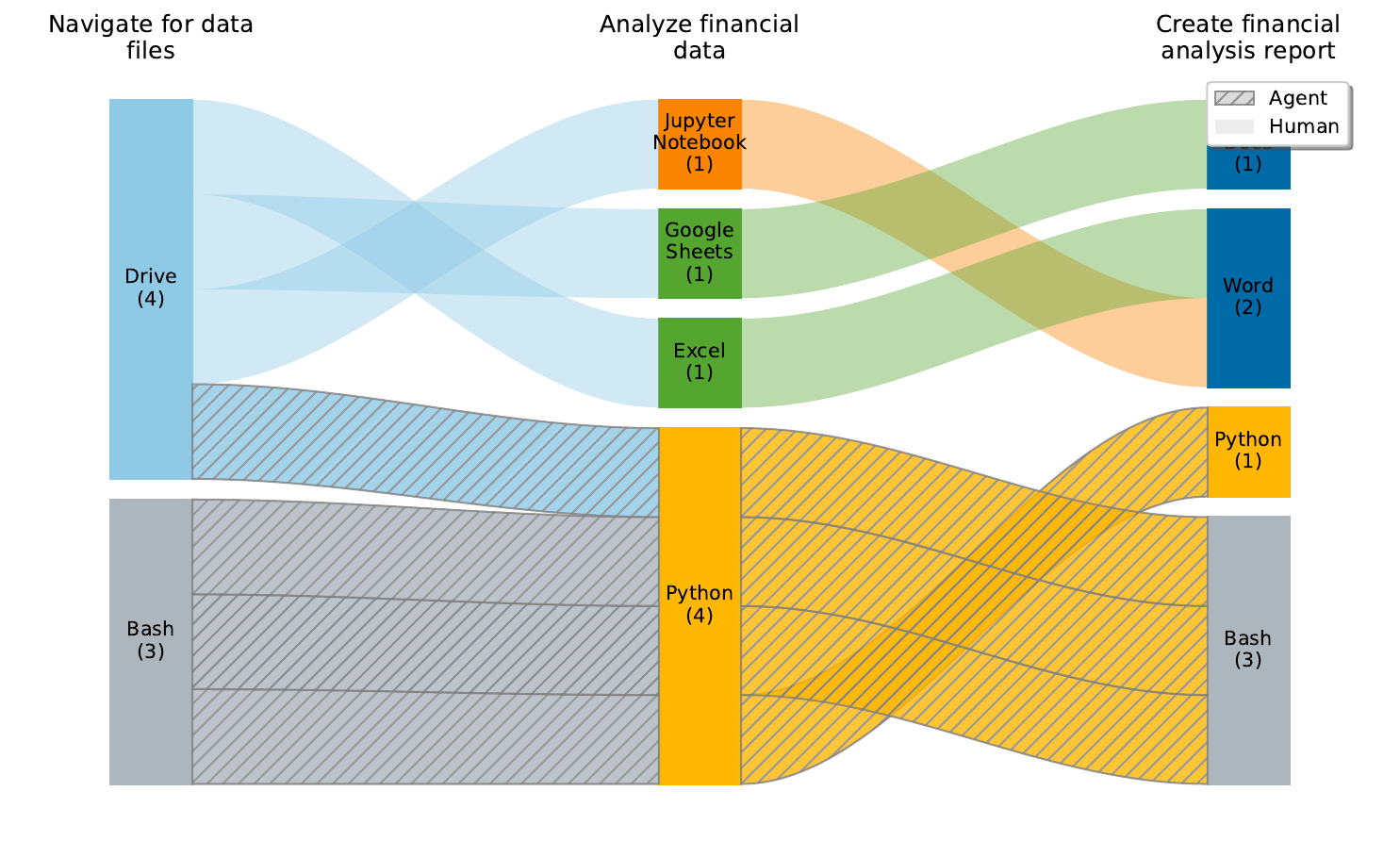}
    \hfill
    \includegraphics[width=0.49\textwidth]{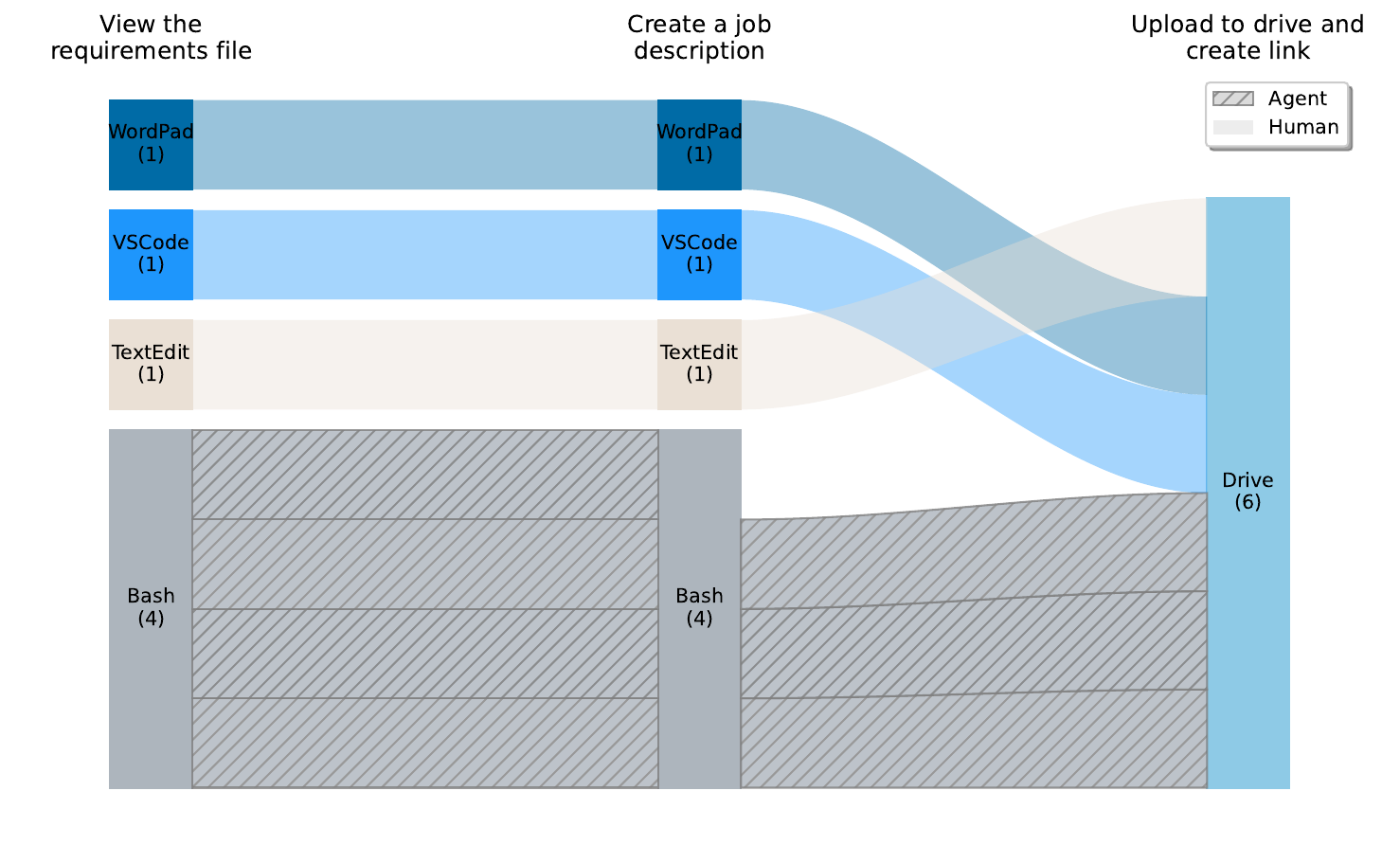}
\caption{\small{Tool usage in \texttt{writing-quarterly-report} (left) and \texttt{writing-new-grad-desc} (right).}}
\label{fig:tool-use-910}
\end{figure*}

\begin{figure*}[ht]
\centering
    \includegraphics[width=0.49\textwidth]{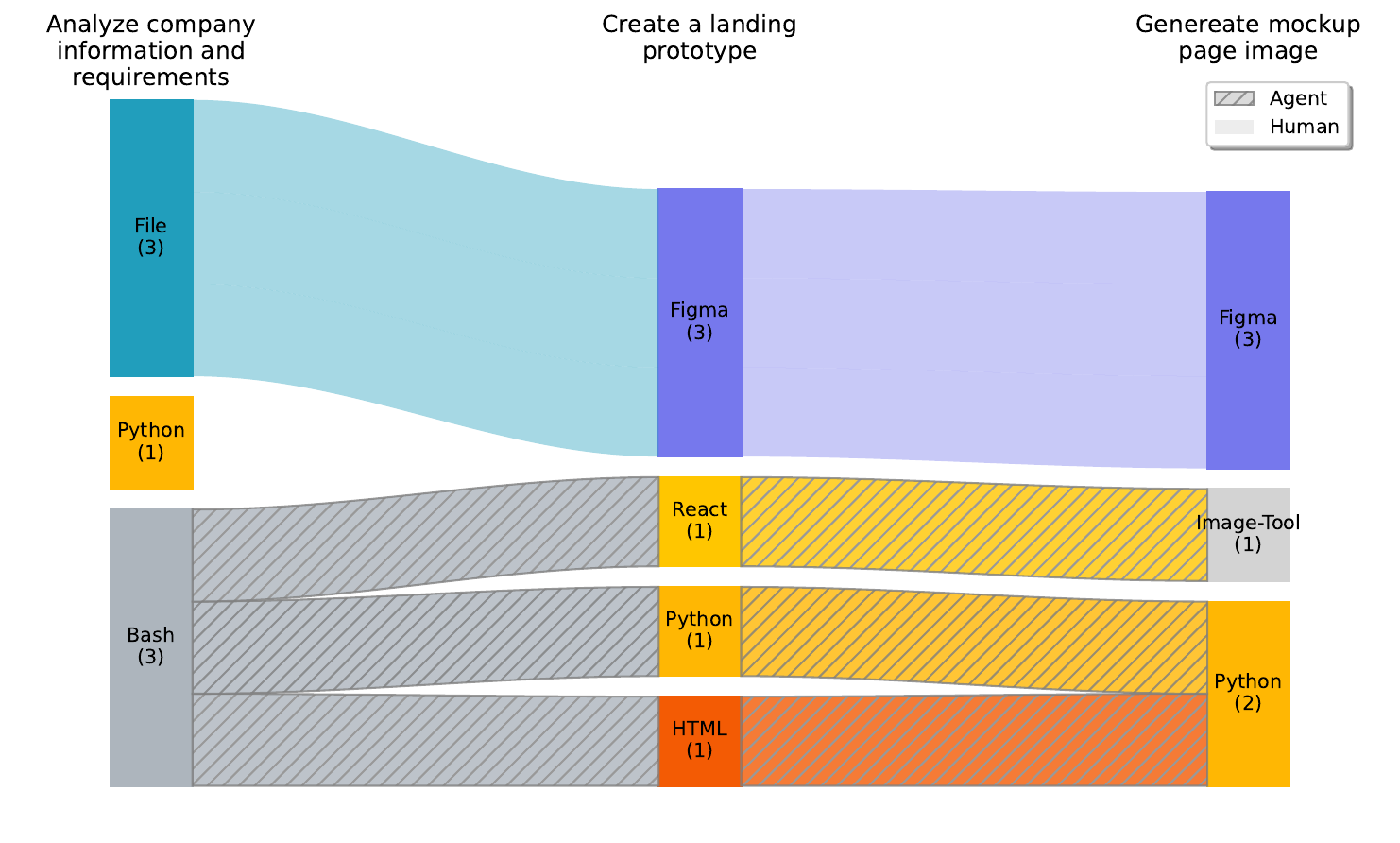}
    \hfill
    \includegraphics[width=0.49\textwidth]{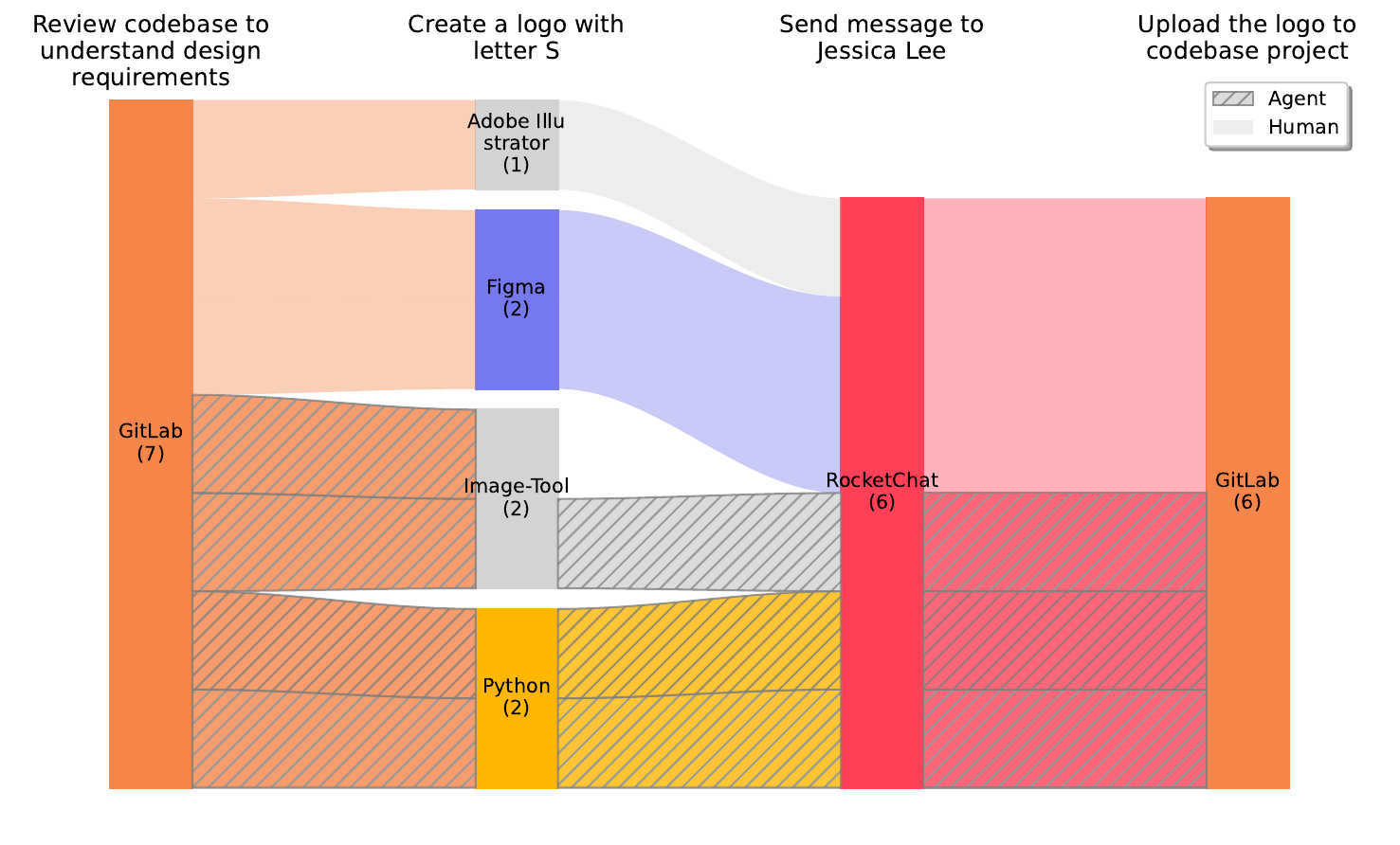}
\caption{\small{Tool usage in \texttt{design-company-landing-page} (left) and \texttt{design-project-logo} (right).}}
\label{fig:tool-use-1112}
\end{figure*}

\clearpage
\section{Agent Work Performance}
\label{app:d:agent-work-perf}

\paragraph{Task Success Rates}
Compared to the progress rate in \autoref{fig:agent-progress}, agents' success rates are 32.5--49.5\% lower. In other words, agents often prioritize making apparent progress, rather than correctly completing, or in some cases even attempting, individual steps. 

\begin{table*}[t!]
\centering
\small
\resizebox{0.98\textwidth}{!}{
\begin{tabular}{l|cc|cccc}
\toprule
\multicolumn{1}{c|}{\multirow{2}{*}{\bf Worker}}  & \multirow{2}{*}{\bf Average Human} & \multirow{2}{*}{\bf Average Agent} & \multicolumn{4}{c}{\bf Agent} \\
{} & {} & {} & {\bf OH, GPT} & {\bf OH, Claude} & {\bf ChatGPT} & {\bf Manus} \\
\midrule
{overall} & {84.6} & {47.3} & {34.5} & {50.3} & {51.1} & {53.0} \\
\midrule
{data analysis} & {82.3} & {52.1} & {34.4} & {40.6} & {61.5} & \markcell{71.9} \\
{engineering}   & {91.7} & {25.0} & {22.9} & {41.7} & {35.4} & {$~~$0.0} \\
{computation}   & {71.5} & {49.3} & {29.5} & {42.0} & \markcell{58.5} & \good{67.4} \\
{writing}       & {94.4} & {64.6} & {50.0} & \good{91.7} & {33.3} & \markcell{83.3} \\
{design}        & {91.7} & {60.6} & {52.5} & {62.5} & {65.0} & {62.5} \\
\bottomrule
\end{tabular}
}
\caption{Agent and human success rates, over all tasks and within each skill category. We mark agent success rates within 10\% and 15\% range of human workers with \small{\colorbox{applegreen!10}{green}} and \small{\colorbox{y!20}{yellow}} colors.}
\label{tab:task-success-rate}
\end{table*}

\begin{figure*}[!t]
\centering
    \includegraphics[width=0.99\textwidth]{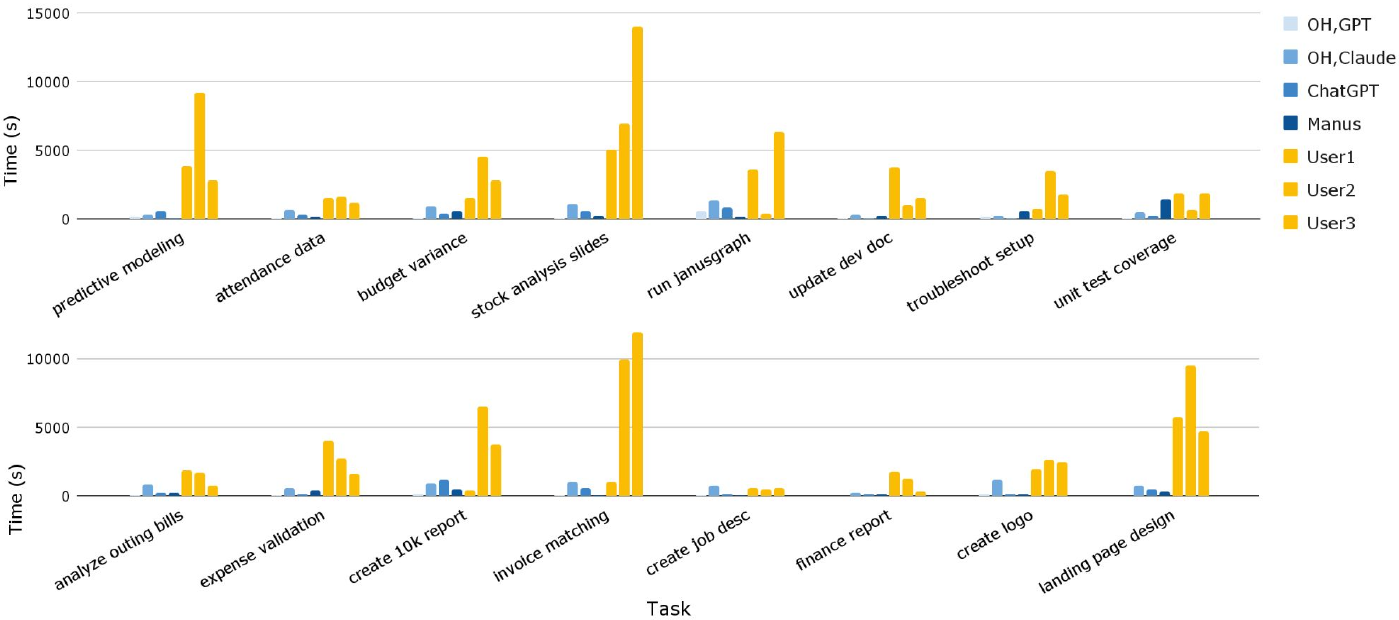}
\caption{Agent workers take substantially less time to finish all tasks.}
\label{fig:work-efficiency}
\end{figure*}

\noindent \textbf{Time Breakdown by Tasks} \quad
In \autoref{fig:work-efficiency}, we present the work time (in seconds) taken by each human and agent worker.

\subsection{Which Domains Do Agents Have Advantages?}
\begin{figure*}[ht]
\vspace{-1mm}
\centering
    \includegraphics[width=0.99\textwidth]{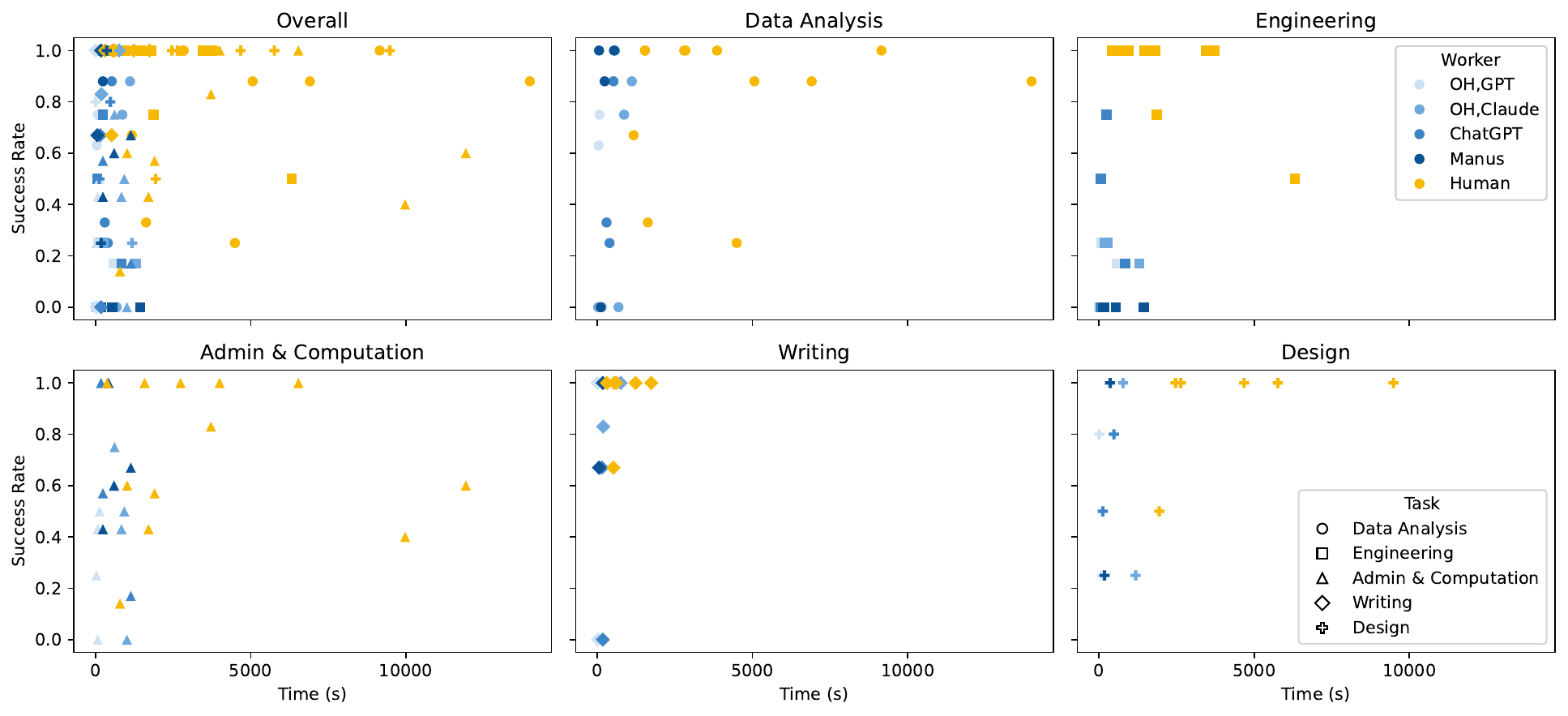}
\vspace{-2mm}
\caption{Time and quality of work from different workers across skill categories.}
\label{fig:time-vs-quality}
\end{figure*}

We compare completion time against work quality in \autoref{fig:time-vs-quality}. 
While agents generally deliver work faster than human workers across most domains, we find that writing and data analysis are the areas where agent-produced work mostly approaches human-level quality. 
Unfortunately, for domains requiring less specialized expertise --- such as administrative and repetitive computational tasks, where AI automation could potentially free up low-wage human labor --- agents only do a mediocre job and are probably not ready for reliable deployment yet \citep{dell2023navigating}. These findings suggest that further training is needed to strengthen fundamental capabilities such as visual understanding and UI interaction.

\noindent $\bullet$ \textbf{Data Analysis} \quad
Agents generally perform well on data analysis tasks; however, the most severe issue is erroneous calculations in 37.5\% of the cases, likely due to false assumptions about instructions, indicating their limited pragmatic reasoning in realistic work contexts. Unfortunately, such errors are difficult to detect and call for more nuanced methodological designs.
Less frequently (12.5\% of cases), agents get stuck at file navigation. As we will show in \S\ref{sec:6.1:human-agent-teaming}, this limitation can be temporarily mitigated by delegating these steps to human workers.

\noindent $\bullet$ \textbf{Engineering} \quad
Despite the widespread development of agents for software engineering, it is somewhat surprising to find agents positioned in the lower-left corner in the visualization, indicating lower quality relative to other skill domains. 
One reason is that our tasks require agents to act in a computer-wise work environment rather than a purely coding environment. For example, agents frequently encounter challenges in configuring runtime environments, executing scripts, or launching servers.
Although the open-source, coding-oriented OpenHands framework performs comparably to general-purpose state-of-the-art agents, the ChatGPT agent struggles more in accessing the sandboxed web environments, likely due to security and authentication constraints.

\noindent $\bullet$ \textbf{Computation} \quad
These tasks, typically considered basic and repetitive for human workers, primarily require basic visual skills (e.g., extracting numbers from bill images) and performing repetitive processes over a long horizon (e.g., thousands of steps).
Unfortunately, agents currently lack robust visual understanding and long-horizon planning abilities, making them ill-suited to automate such work.
This finding is somewhat disappointing, as these lower-entry-bar roles are among those most desirable for early automation. Instead, agents appear more capable in expert-oriented domains such as engineers \citep{cui2025effects}, suggesting that these routine computational jobs may remain human-dominated for some time.

\noindent $\bullet$ \textbf{Writing} \quad
While \autoref{fig:time-vs-quality} shows that agents are the closest at automating writing jobs currently done by humans, this may be partly due to the structured nature of the writing tasks in our study --- HR drafting job descriptions or analysts composing financial reports with predefined modules. In contrast, many other forms of writing (e.g., books, articles) are considerably more open-ended and creative. Nevertheless, when viewed through the lens of writing as a skill rather than writing as a profession, many occupational writing activities (e.g., by engineers or analysts) tend to follow structured patterns. As prior studies in domain-specific writing, such as legal \citep{choi2024ai} and academic \citep{lee2022coauthor,zhao2025knoll} contexts, report varying findings, our findings should be interpreted with caution and contextual awareness.

\begin{wrapfigure}[10]{r}{0.2\textwidth}
\vspace{-3mm}    
\includegraphics[width=0.2\textwidth]{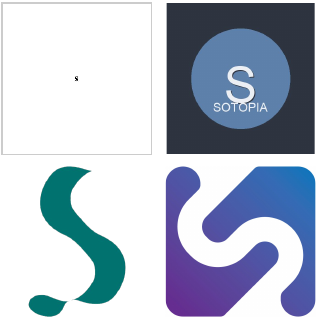}
\vspace{-5mm}
\caption{Agent (top) and human (bottom) designs.}
\label{fig:agent-vs-human-design}
\end{wrapfigure}

\noindent $\bullet$ \textbf{Design} \quad
Agent designers display patterns similar to those of agent administrative staff --- producing work of moderate quality (\autoref{fig:agent-vs-human-design}) but with notable efficiency advantages. They may thus be useful in contexts where quality requirements are not stringent, or as effective starting points that human designers can later refine.

\subsection{Task Delegation by Programmability}
\label{app:d.2:task-programmability}
Given that agents are better at solving tasks via programmatic approaches, while humans excel at operating through UI interfaces, an important question arises: what types of work are best suited to each?
Some tasks, or specific workflow steps within a task, can be readily programmed, whereas others cannot. We hence propose three levels of task programmability and discuss their respective implications for agent and human workers.

\noindent $\bullet$ \textbf{Readily Programmable}: These tasks can often be solved reliably through deterministic program execution. For example, writing a Python script to clean an Excel datasheet or coding a website in HTML. Compared with the UI-oriented tools preferred by many human workers, programmatic approaches often offer a stronger guarantee of outcome accuracy and greater scalability of the process. For instance, processing 10000 data entries in Python is typically faster and less error-prone than manually editing a 10$k$-row Excel sheet. 
While some human workers have begun integrating programming into their workflows, many still rely on UI tools and consequently operate less efficiently. Given current advances in agent programming capabilities, such tasks are likely the most suitable for agent automation at present.

\noindent $\bullet$ \textbf{Half Programmable}: Some tasks are theoretically programmable, but lack clear, direct programmatic paths within the same tool used by human workers. 
For instance, to generate a \texttt{docx} report, agents usually write the content in \texttt{Markdown} format and then convert it to \texttt{docx} files using auxiliary programming libraries. Similarly, instead of operating through \texttt{Figma}'s interface, agents may write \texttt{HTML} code to produce a comparable website design, albeit through a different underlying logic.

\begin{wrapfigure}[15]{r}{0.6\textwidth}
\vspace{-5mm}    
\includegraphics[width=0.6\textwidth]{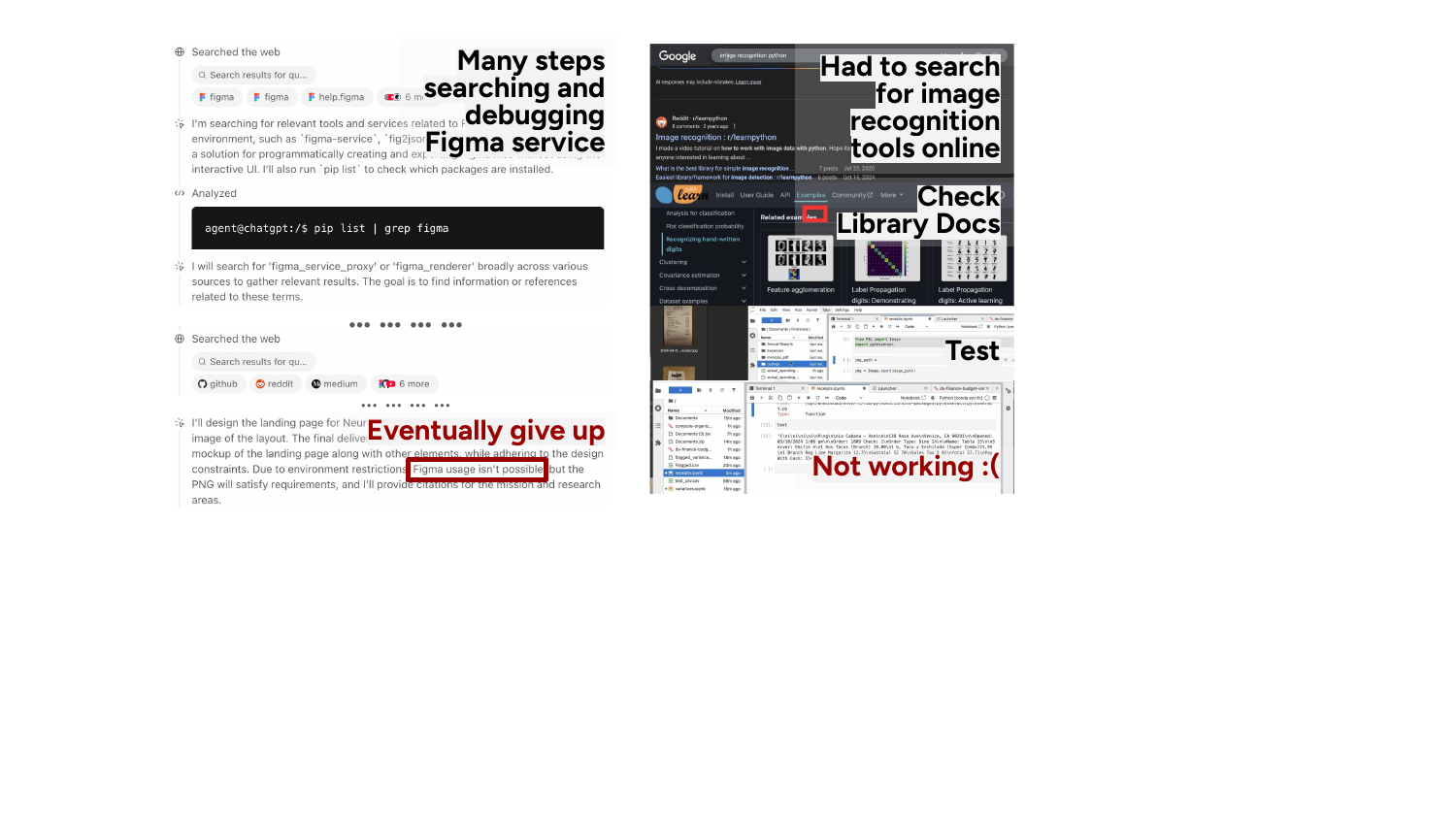}
\vspace{-4mm}
\caption{Left: Agent attempts to use Figma tool for design but fails. Right: Human attempts to use program tool for bill processing but fails.}
\label{fig:the-agent-human-way}
\end{wrapfigure}

For this type of task, it is not entirely clear whether progress should focus on advancing the agent's programmatic approach or on better emulating the human workflow. For humans, who typically rely on visually oriented tools for such tasks, it is often less intuitive to conceptualize how these operations could be performed programmatically. 
Agents, on the other hand, struggle with UI interaction: in the ChatGPT design process depicted in \autoref{fig:the-agent-human-way} (left), when attempting to use the Figma design interface, it spends numerous steps figuring out the tool before eventually reverting to writing programs --- showing its limitations in interacting with professional UI environments. 
Beyond advancing agents' basic UI skills, further engineering efforts could focus on exposing or extending API-based programming pathways, such as supporting \texttt{figma} API accesses or developing alternative programming paths with equivalent functionalities.

\noindent $\bullet$ \textbf{Less Programmable}: Some tasks rely heavily on visual perception and lack a deterministic programmatic solution. For instance, in the task of viewing bill images and extracting data from them (\autoref{fig:the-agent-human-way}, right), a programmatic solution may require non-deterministic modules, such as invoking a neural OCR model to parse out the text, which may not guarantee task success \citep{wang2024what}. 
Even human engineers, who can easily script some tasks such as data analysis, encounter challenges in this case. As in \autoref{fig:the-agent-human-way} (right), the human worker's programmatic attempt fails after searching for relevant OCR packages, observing undesired parsing performance, and ultimately abandoning the task.
For such cases, it is often far easier for humans to simply inspect the images and manually enter the numbers in an Excel sheet via the UI.
Although these tasks may be long-tail, particularly after more tasks become programmatically solvable, they remain inevitable components of computer-use activities. Many of our analysis provides substantial evidence of agents' limitations in these less-programmable scenarios, underscoring the need for continued improvement through foundation model training.

\end{document}